%% file: iros_rewrite_cameraready_arxiv.tex
\pdfoutput=1
\documentclass[letterpaper, 10 pt, conference]{ieeeconf}
\IEEEoverridecommandlockouts     

\usepackage{lipsum}

\usepackage{multibib}
\newcites{app}{Appendix References}

\usepackage{wrapfig}
\usepackage{xcolor} 

\usepackage{amsmath}
\usepackage{amsfonts}
\usepackage{amsthm}
\usepackage{amssymb}
\usepackage{mathrsfs}  
\usepackage{bm}
\usepackage[ruled,vlined]{algorithm2e}
\usepackage{mathdots} 
\usepackage{dsfont} 
\usepackage{accents} 
\usepackage{tikz-cd} 
\usepackage{adjustbox} 
\usepackage{mathtools} 

\usepackage{caption}
\usepackage{multirow}
\usepackage{multicol}
\usepackage{mdframed} 

\usepackage{listings}
\usepackage{graphicx}

\usepackage{balance}

\newcommand{\comment}[1]{} 
 
\newcommand{\norm}[1]{\left\lVert#1\right\rVert} 
\newcommand\mat[1]{\begin{bmatrix}#1\end{bmatrix}} 
\newcommand\eqs[1]{\begin{equation}\begin{split}#1\end{split}\end{equation}} 
\newcommand\eqsnn[1]{\begin{equation*}\begin{split}#1\end{split}\end{equation*}} 
\DeclareMathOperator*{\setint}{int} 

\DeclareMathOperator*{\diag}{diag} 

\DeclareMathOperator*{\conv}{conv} 
\DeclareMathOperator*{\minimize}{minimize} 
\DeclareMathOperator*{\maximize}{maximize}
\DeclareMathOperator*{\find}{find}
\DeclareMathOperator*{\subto}{subject\;to}
\DeclareMathOperator*{\erf}{erf} 
\newcommand\braces[1]{\left\{#1\right\}} 
\newcommand\brackets[1]{\left[#1\right]} 
\newcommand\parens[1]{\left(#1\right)} 
\newcommand\abs[1]{\left|#1\right|} 
\newcommand\wedgeop[1]{\left[#1\right]_\times}
\newcommand\R{\mathbb{R}} 

\input{notation}

\newtheorem{corollary}{Corollary}
\newtheorem{theorem}{Theorem}
\newtheorem{lemma}{Lemma}

\newtheorem{definition}{Definition}
\newtheorem*{definition*}{Definition}

\newtheorem{proposition}{Proposition}

\usepackage[bookmarks=true, hidelinks]{hyperref}

\makeatletter
\let\save@mathaccent\mathaccent
\newcommand*\if@single[3]{%
  \setbox0\hbox{${\mathaccent"0362{#1}}^H$}%
  \setbox2\hbox{${\mathaccent"0362{\kern0pt#1}}^H$}%
  \ifdim\ht0=\ht2 #3\else #2\fi
  }
\newcommand*\rel@kern[1]{\kern#1\dimexpr\macc@kerna}
\newcommand*\widebar[1]{\@ifnextchar^{{\wide@bar{#1}{0}}}{\wide@bar{#1}{1}}}
\newcommand*\wide@bar[2]{\if@single{#1}{\wide@bar@{#1}{#2}{1}}{\wide@bar@{#1}{#2}{2}}}
\newcommand*\wide@bar@[3]{%
  \begingroup
  \def\mathaccent##1##2{%
    \let\mathaccent\save@mathaccent
    \if#32 \let\macc@nucleus\first@char \fi
    \setbox\z@\hbox{$\macc@style{\macc@nucleus}_{}$}%
    \setbox\tw@\hbox{$\macc@style{\macc@nucleus}{}_{}$}%
    \dimen@\wd\tw@
    \advance\dimen@-\wd\z@
    \divide\dimen@ 3
    \@tempdima\wd\tw@
    \advance\@tempdima-\scriptspace
    \divide\@tempdima 10
    \advance\dimen@-\@tempdima
    \ifdim\dimen@>\z@ \dimen@0pt\fi
    \rel@kern{0.6}\kern-\dimen@
    \if#31
      \overline{\rel@kern{-0.6}\kern\dimen@\macc@nucleus\rel@kern{0.4}\kern\dimen@}%
      \advance\dimen@0.4\dimexpr\macc@kerna
      \let\final@kern#2%
      \ifdim\dimen@<\z@ \let\final@kern1\fi
      \if\final@kern1 \kern-\dimen@\fi
    \else
      \overline{\rel@kern{-0.6}\kern\dimen@#1}%
    \fi
  }%
  \macc@depth\@ne
  \let\math@bgroup\@empty \let\math@egroup\macc@set@skewchar
  \mathsurround\z@ \frozen@everymath{\mathgroup\macc@group\relax}%
  \macc@set@skewchar\relax
  \let\mathaccentV\macc@nested@a
  \if#31
    \macc@nested@a\relax111{#1}%
  \else
    \def\gobble@till@marker##1\endmarker{}%
    \futurelet\first@char\gobble@till@marker#1\endmarker
    \ifcat\noexpand\first@char A\else
      \def\first@char{}%
    \fi
    \macc@nested@a\relax111{\first@char}%
  \fi
  \endgroup
}
\makeatother

\begin{document}

\title{
    Toward An Analytic Theory of Intrinsic Robustness for Dexterous Grasping
}

\author{
    Albert H. Li$^\dagger$, Preston Culbertson$^\ddagger$, Aaron D. Ames$^{\dagger,\ddagger}$%
    \thanks{$\dagger$ A. H. Li and A. D. Ames are with the Department of Computing and Mathematical Sciences, California Institute of Technology, Pasadena, CA 91125, USA, \texttt{\{alberthli, ames\}@caltech.edu}.}%
    \thanks{$\ddagger$ P. Culbertson and A. D. Ames are with the Department of Civil and Mechanical Engineering, California Institute of Technology, Pasadena, CA 91125, USA, \texttt{\{pculbert, ames\}@caltech.edu}.}%
}

\maketitle

\begin{abstract}
Conventional approaches to grasp planning require perfect knowledge of an object's pose and geometry. Uncertainties in these quantities induce uncertainties in the quality of planned grasps, which can lead to failure. Classically, grasp robustness refers to the ability to resist external disturbances \textit{after} grasping an object. In contrast, this work studies robustness to intrinsic sources of uncertainty like object pose or geometry affecting grasp planning \textit{before} execution. To do so, we develop a novel analytic theory of grasping that reasons about this \textit{intrinsic robustness} by characterizing the effect of friction cone uncertainty on a grasp's force closure status. We apply this result in two ways. First, we analyze the theoretical guarantees on intrinsic robustness of two grasp metrics in the literature, the classical Ferrari-Canny metric and more recent min-weight metric. We validate these results with hardware trials that compare grasps synthesized with and without robustness guarantees, showing a clear improvement in success rates. Second, we use our theory to develop a novel analytic notion of \textit{probabilistic} force closure, which we show can generate unique, uncertainty-aware grasps in simulation.
\end{abstract}

\maketitle

\section{Introduction}\label{sec:intro}
Grasp synthesis has been a canonical problem in robotic manipulation since the field's inception. Despite decades of work, dexterous grasping is still challenging, since multi-finger hands are high-dimensional and have complex kinematics. Broadly, two approaches toward dexterous grasping exist.
\textit{Analytic} methods maximize a grasp metric that measures grasp quality \cite{roa2014_graspmetricssurvey}. Such methods are principled and intuitive, but suffer from two main drawbacks: (a) they usually require perfect knowledge of an object, and (b) efficiently optimizing metrics is typically hard while enforcing kinematic and collision constraints \cite{li1988_taskorientedgrasping}.
In response, many \textit{data-driven} methods for grasping account for uncertainty implicitly using empirical grasp data, but lack guarantees or interpretability, often checking constraint satisfaction \textit{post hoc} rather than enforcing them during synthesis \cite{kappler2015_bigdatagrasping, shao2019_unigrasp}. Since high-quality, feasible grasps are rare to find randomly, these methods usually require many grasp data for training, which may be difficult to acquire (particularly from hardware).

\begin{figure}
    \centering
    \includegraphics[width=\linewidth]{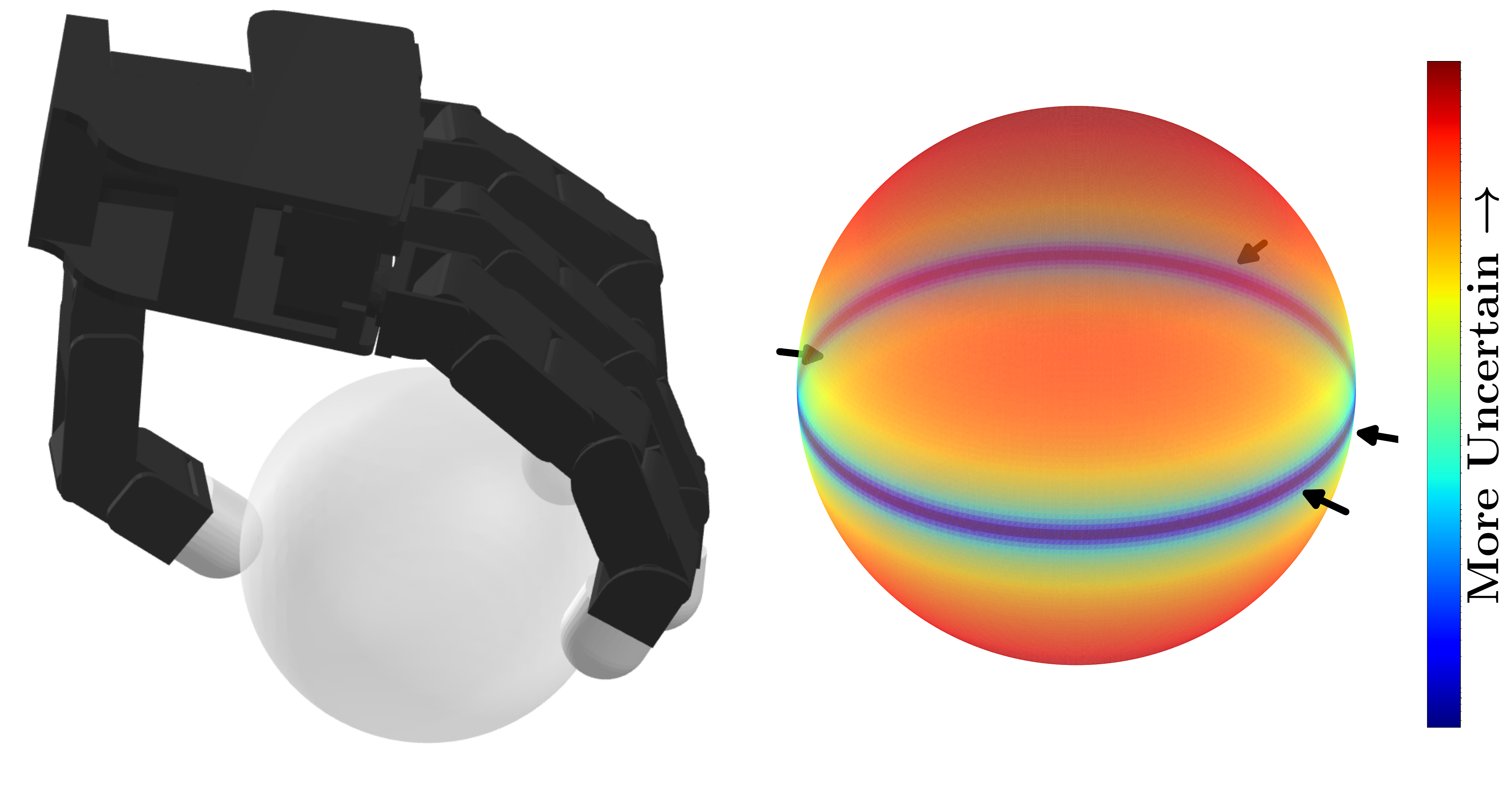}
    \caption{This work develops an analytic theory of uncertainty-aware grasping, which enables an analysis of the uncertainty tolerance of classical grasp metrics, and the development of new, probabilistic approaches. Above we plot a grasp obtained by optimizing our proposed probabilistic metric, PONG, on a toy example with varying surface normal uncertainty. While there exist many robust grasps for the true geometry, the grasp synthesized using our probabilistic metric places the fingertips in the minimum-uncertainty region.}
    \label{fig:banner}
    \vspace{-0.5cm}
\end{figure}

Classically, grasp robustness reports the size of \textit{extrinsic} disturbances a grasp can resist post-execution, like an unexpected force on the object \cite{ferraricanny1992}. But, many uncertainties we call \textit{intrinsic} (like those in object pose or geometry) affect planning \textit{before} execution. Poor plans create brittle grasps, so intrinsic uncertainties reduce extrinsic robustness. Despite this, intrinsic robustness remains understudied, and the aim of this work is to provide new theory to analyze it.

Our main contribution is a mathematical sufficient condition that guarantees intrinsic grasp robustness under a linear friction cone model. This immediately proves that the classical Ferrari-Canny metric $\varepsilon$ \cite{ferraricanny1992}, designed to characterize resistance to extrinsic disturbances, also measures intrinsic robustness (Theorem \ref{thm:containment} and Corollary \ref{cor:fc_uncertainty_aware}). We further apply our result in two ways to demonstrate its utility.

First, we prove the recently proposed \textit{min-weight metric} $\ell^*$, computable via solving an LP, \cite{li2023_frogger} uniformly lower bounds (a positive multiple of) the Ferrari-Canny metric $\varepsilon$, so $\ell^*$ grants intrinsic robustness and maximizing it is justified. We validate the theory by comparing $\ell^*$-synthesized grasps against a competitive baseline on dozens of hardware trials, showing superior performance.

Second, we develop a novel \textit{analytic} notion of probabilistic force closure, and develop an uncertainty-aware metric called \textit{probabilistic object normals for grasping} (PONG). We demonstrate that maximizing PONG on objects with synthetic belief distributions leads to probabilistically-robust grasps (e.g., Fig. \ref{fig:banner}). In simulation, we show that $\ell^*$ performs similarly to PONG on uncertain objects, showing that the aforementioned bound is not conservative in practice.

\subsection{Related Work}\label{sec:related_work}
While this paper presents an analytic treatment of intrinsic uncertainty, prior work has studied it from data-driven perspectives. For example, \cite{weisz2012_pfc_pose_robust} notes that $\varepsilon$ is sensitive to uncertainty in object pose and estimates force closure probability via Monte Carlo simulation to rank grasps by robustness. \textit{Gaussian process implicit surfaces} (GPISs) \cite{dragiev2011_gpis} have represented uncertain geometry for parallel-jaw grasping \cite{mahler2015_gpis}, dexterous grasping with tactile sensors \cite{defarias_tactile_gpis_uncertain}, and control synthesis \cite{li2016_shape_uncertainty}. But, GPISs are difficult to supervise from sensor data (requiring known signed-distance labels), scale poorly with data quantity, and lack expressivity (due to strong smoothness priors imposed by usual kernels). 

An alternate approach is to directly learn a probabilistic metric from data. For example, Dex-Net 2.0 predicts the robustness of a batch of planar parallel-jaw grasps, selecting the most robust one \cite{mahler2017_dexnet2}. Similar works have optimized multi-finger grasps with gradient-based optimization by leveraging the differentiability of neural networks \cite{lu2018_planningmultifinger, lu2020_diffgrasplearning}.

Many works have proposed differentiable approximations of analytic metrics or the force closure condition for gradient-based grasp synthesis. Such methods include solving a large sum of squares program \cite{liu2020_deepdiffgrasp}; optimizing a penalty-based relaxation of the force closure condition \cite{liu2021_diversediffgrasps, wang2022_dexgraspnet}; solving a bilinear optimization program with a QP force closure constraint \cite{wu2022_learningdexgraspsgenmodel}; and maximizing a differentiable proxy for the Ferrari-Canny metric called the min-weight metric $\ell^*$ \cite{li2023_frogger}. However, it is typically not understood how well or whether these approximations preserve grasp robustness.

We use our results to prove that maximizing $\ell^*$, the fastest of the above methods, equivalently maximizes intrinsic robustness, which theoretically justifies its use as a fast and principled approximation of the Ferrari-Canny metric. We note that while other works have derived lower bounds for $\varepsilon$ (e.g., \cite{pokorny2013_lowerboundfc, dai2015_forceclosuresdp, liu2020_deepdiffgrasp}), the cost of computing these metrics is significantly higher than for $\ell^*$, which requires only solving a small linear program. In future work, we hope to similar analyze the robustness properties of other metrics.

\subsection{Mathematical Preliminaries}\label{sec:preliminaries}
We consider a fixed-base, fully-actuated rigid-body serial manipulator with a multi-finger hand and configuration $q\in\mathcal{Q}$. Assume the hand has $n_f$ fingers contacting the object at points $\{\pos^i\}_{i=1}^{n_f}$ with inward pointing surface normals $\{n^i\}_{i=1}^{n_f}$ and forward kinematics maps $\pos^i=FK^i(q)$. Denote the compact manipuland $\mathcal{O}\subset\R^3$ with surface $\partial \mathcal{O}$.

As shorthand, let $\mathcal{C}_x:=\conv(\{x_l\})$ denote the convex hull of a finite set of points indexed by $l$. Define the \textit{wedge} operator $\wedgeop{\cdot}: \R^3 \rightarrow \mathfrak{so}(3)$ such that $\wedgeop{a} b := a \times b$ for $a,b\in\R^3$. We denote an optimized \textit{feasible} configuration as $q^*$, meaning no undesired collisions while contacting the object. The fingers are modeled as point contacts with friction.

Recall that for Coulomb friction, a contact force $f$ satisfies the \textit{no-slip} condition if for friction coefficient $\mu$, $\norm{f_t}\leq\mu\cdot f_n$, where $t$ and $n$ denote tangent and (positive) normal components of $f$. We call such forces \textit{Coulomb-compliant}. A force applied at point $x$ induces the torque $\tau=x \times f$, which in turn induces a corresponding wrench $w=(f, \tau)$.

The \textit{friction cone} at a point $x$ is the cone centered about a surface normal $n$ consisting of all Coulomb-compliant forces. We use a standard pyramidal approximation of it with $n_s$ sides, calling its edges \textit{basis forces}. We emphasize that the induced \textit{basis wrenches} depend entirely on these pyramids via the basis forces, which depend on a normal $n$ at a point $x$; so, we always implicitly have $w=w(q)$ and we refer to basis wrench uncertainty equivalently as \textit{friction cone uncertainty}. Since it depends on $x$ and $n$, the friction cone implicitly captures the effects of intrinsic uncertainty on grasps.

We let $n_w=n_fn_s$ denote the number of basis wrenches (indexed $w_j^i$), each associated with finger $i$ and pyramid edge $j$. For brevity, we often combine indices $i,j$ into a single one $l$. Let $\mathcal{I}=\{1,\dots,n_f\}, \mathcal{J}=\{1,\dots,n_s\}, \mathcal{L}=\{1,\dots,n_w\}$ denote index sets, and $\mathcal{W}=\{w_l\}_{l=1}^{n_w}$ denote a set of basis wrenches. Optimal solutions to optimization programs are denoted with asterisks. A set is \textit{degenerate} if it has zero volume. $B_r(c)$ denotes a ball of radius $r$ centered at $c$.

\begin{figure*}
    \centering
    \includegraphics[width=\linewidth, trim={0.7cm 0 1cm 0}, clip]{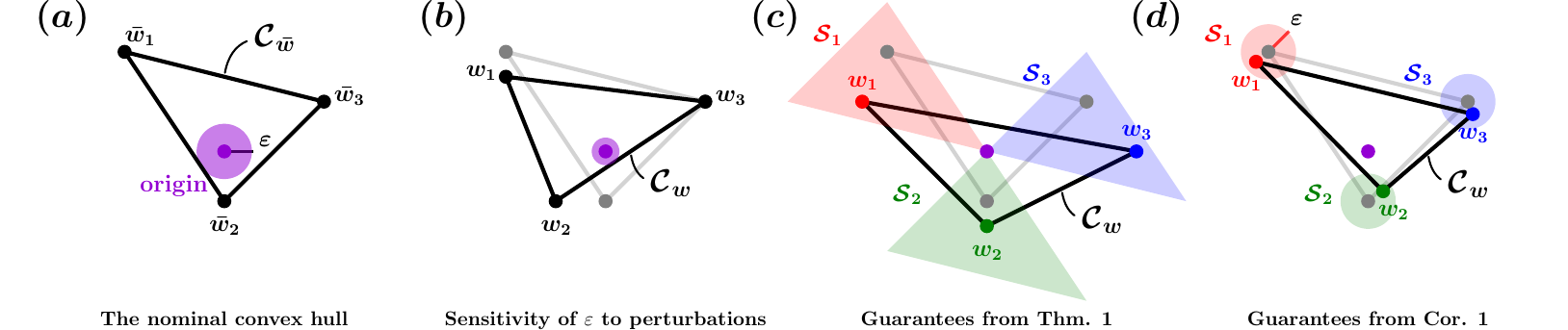}
    \caption{
    \textbf{(a)} A set of nominal basis wrenches and their hull $\mathcal{C}_{\bar{w}}$ with a Ferrari-Canny metric of $\varepsilon$. \textbf{(b)} The value of $\varepsilon$ is sensitive to some perturbations (e.g., $w_2$) but not others (e.g., $w_1$). We derive guarantees on the size of allowable perturbations such that $0\in\mathcal{C}_w$. \textbf{(c)} The guarantees from Theorem \ref{thm:containment}. The shaded regions $\mathcal{S}_i$ indicate areas where $w_i$ may lie, independent of other basis wrenches, while still guaranteeing $0\in\mathcal{C}_w$. Note how the $\mathcal{S}_i$ are anisotropic. \textbf{(d)} The Ferrari-Canny metric $\varepsilon$ also provides (strictly worse) guarantees as a result. Note this cartoon shows a planar wrench space, which is actually $\R^6$.
    }
    \label{fig:containment}
    \vspace{-0.5cm}
\end{figure*}

\section{A Theory of Intrinsically-Robust Grasping}\label{sec:theory}
Recall that a grasp is force closure if it can generate Coulomb-compliant forces on the object to resist any disturbance wrench. A sufficient condition for force closure is that the origin is contained in the convex hull of a grasp's basis wrenches, i.e., $0\in\mathcal{C}_w$ \cite{li2023_frogger}. We say that such basis wrenches \textit{certify} force closure.

Since the basis wrenches depend on contacts $x$ and surface normals $n$, force closure is a function of both an object's geometry and a grasp's contacts. Thus, intrinsic uncertainties generate uncertainty in a grasp's basis wrenches, also reducing extrinsic robustness. In this section, we analyze how much uncertainty a nominally force closure can tolerate while still remaining force closure.

\subsection{Characterizing a Grasp's Uncertainty Tolerance}
Consider a nominal set of basis wrenches $\mathcal{C}_{\bar{w}}$ and the convex hull of a perturbed version, $\mathcal{C}_w$ (see Fig. \ref{fig:containment}a/b). For a given grasp, the value of $\varepsilon$ may be sensitive or insensitive to such perturbations. If the perturbations are ``large,'' the origin will leave $\mathcal{C}_w$, so we cannot certify force closure. Let $w_l$ and $\bar{w}_l$ denote elements of $\mathcal{C}_w$ and $\mathcal{C}_{\bar{w}}$ respectively. The following result bounds the size of tolerable perturbations in terms of $\mathcal{C}_{\bar{w}}$.

\begin{theorem}\label{thm:containment}
    If $w_l-\bar{w}_l\in -\mathcal{C}_{\bar{w}}$ for all $l\in\mathcal{L}$, then $0\in\mathcal{C}_w$.
\end{theorem}
\begin{proof}
    Suppose for the sake of contradiction that $0\not\in\mathcal{C}_w$. Then, there exists $a$ such that $a^\top w>0$ for all $w\in\mathcal{C}_w$ by the separating hyperplane theorem. Further, there must exist some $k\in\mathcal{L}$ satisfying $a^\top\bar{w}_{k}\leq a^\top\bar{w}_{l}$ for all $l\in\mathcal{L}$.
    
    Since $w_l-\bar{w}_l\in-\mathcal{C}_{\bar{w}}$ for each $l$, there exists a set of convex weights $\alpha\in\R^{n_w}$ (with $\alpha \succeq 0, \sum_{l=1}^{n_w} \alpha_l = 1$) such that $w_{k} - \bar{w}_{k} = -\sum_{l=1}^{n_w}\alpha_l\bar{w}_l.$ Thus,
    \eqs{
        w_{k} = \bar{w}_{k} &+ (w_{k} - \bar{w}_{k}) = \bar{w}_{k}-\sum_{l=1}^{n_w}\alpha_l\bar{w}_l,
    }
    which implies
    \eqs{\label{eqn:lemma_eqn}
        a^\top w_{k} &= a^\top \bar{w}_{k} - \sum_{l=1}^{n_w}\alpha_l a^\top \bar{w}_l \\
        &\leq a^\top \bar{w}_{k} - a^\top \bar{w}_{k}\sum_{l=1}^{n_w}\alpha_l = 0.
    }
    But, $a^\top w_{k}>0$ since $w_{k}\in\mathcal{C}_w$, contradicting \eqref{eqn:lemma_eqn}.
\end{proof}
We view $w_l-\bar{w}_l$ as the deviation of the $l^{th}$ basis wrench from some nominal value. If all deviations are respectively contained in $-\mathcal{C}_{\bar{w}}$, then by Theorem \ref{thm:containment}, the origin lies in the convex hull $\mathcal{C}_w$ of the true, perturbed basis wrenches, i.e., \textit{the grasp is force closure}. See Fig. \ref{fig:containment}c for visual intuition.

While force closure is a binary measure of grasp quality, the Ferrari-Canny metric \cite{ferraricanny1992} reports the (extrinsic) \textit{robustness} of force closure grasps by computing the radius $\varepsilon$ of the largest origin-centered ball contained in $\mathcal{C}_w$. The grasp can resist any applied disturbance wrench with a norm smaller than $\varepsilon$, with the type of norm depending on assumed constraints on a hand's control authority \cite{ferraricanny1992}.

\begin{corollary}\label{cor:fc_uncertainty_aware}
    Let $\varepsilon$ denote the Ferrari-Canny metric for a nominal set of basis wrenches $\widebar{\mathcal{W}}$. If the true (i.e., perturbed) basis wrenches $w_1,\dots,w_{n_w}$ satisfy
    \eqs{
        w_l \in B_{\varepsilon}(\bar{w}_l),\;\forall l \in \mathcal{L},
    }
    then the true grasp is force closure.
\end{corollary}
\begin{proof}
    Follows by Theorem \ref{thm:containment}, since $B_\varepsilon(0) \subset -\mathcal{C}_{\bar{w}}$. Therefore, $0\in\mathcal{C}_w$, so the true grasp is force closure.
\end{proof}
Though $\varepsilon$ represents a grasp's sensitivity to external disturbance wrenches, Corollary \ref{cor:fc_uncertainty_aware} shows that $\varepsilon$ also captures a grasp's \textit{intrinsic} robustness (e.g., to object pose/geometry) implicitly through friction cone uncertainty. We analyze concrete ways to explicitly use the relation between object geometry and basis wrenches in Sec. \ref{sec:pong}. See Fig. \ref{fig:containment}d for an illustration of Corollary \ref{cor:fc_uncertainty_aware}.

\begin{figure*}[htbp]
    \centering
    \includegraphics[width=\linewidth]{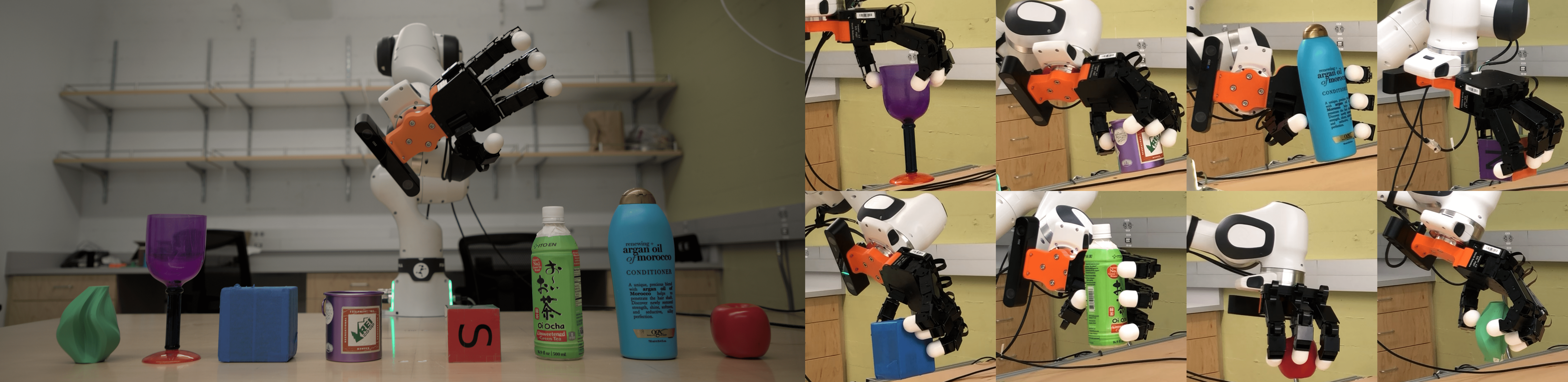}
    \caption{\textbf{Left.} Hardware setup. We synthesized grasps on eight objects for an Allegro hand mounted on a Franka Research 3. From left to right: 3D-printed part, goblet, box, mug, cube, bottle, conditioner, and apple. We employed an eye-in-hand setup with a Zed camera to capture images used to train a NeRF prior to grasping. Only monocular RGB was used. \textbf{Right.} Representative FRoGGeR-synthesized grasps.}
    \label{fig:hardware_setup}
    \vspace{-0.5cm}
\end{figure*}

\subsection{\texorpdfstring{Bounding $\varepsilon$ with the Min-Weight Metric $\ell^*$}{Bounding eps with the Min-Weight Metric l*}}
One challenge for grasp synthesis with the Ferrari-Canny metric $\varepsilon$ is that it is slow to optimize, especially due to its non-differentiability \cite{li2023_frogger}. This motivates efficient approximations of $\varepsilon$ like the \textit{min-weight metric} $\ell^*$, an almost-everywhere differentiable proxy for the Ferrari-Canny metric computable by solving a simple LP. In this section, we leverage Corollary \ref{cor:fc_uncertainty_aware} (and by extension, Theorem \ref{thm:containment}) to demonstrate that maximizing $\ell^*$ provides guarantees on intrinsic robustness. The key heuristic of the min-weight metric is to encourage the origin to lie far inside $\mathcal{C}_w$ by maximizing its minimum convex weight (see \cite{li2023_frogger}).

The following LP defines $\ell^*$:
\eqs{\label{eqn:min_weight_primal}
    \ell^* = \maximize_{\alpha \in \R^{n_w},\ell \in \R}\quad & \ell \\
    \textrm{subject to} \quad & W\alpha = 0 \\
    & \mathds{1}_{n_w}^\top \alpha = 1 \\
    & \alpha \succeq \ell\mathds{1}_{n_w},
}
where $W=W(q)$ is the matrix whose $l^{th}$ column is $w_l\in\mathcal{W}$ and $\mathds{1}_{n_w}\in\R^{n_w}$ is the vector of all 1s.

In \cite[Fig. 3]{li2023_frogger}, it was observed that when empirically evaluated on force closure grasps, (an unknown positive scaling $K$ of) $\ell^*$ seemed to lower bound $\varepsilon$ uniformly. This motivated the maximization of $\ell^*$ as a conservative approximation of $\varepsilon$. If true, this would imply that $\ell^*$ captures intrinsic robustness by Corollary \ref{cor:fc_uncertainty_aware}, since $B_{K\ell^*}(0) \subseteq B_{\varepsilon}(0)$. In this section, we formally prove the claim for ``sufficiently regular'' grasps.

Every compact set $\mathcal{X}$ has a largest (not necessarily origin-centered) inscribing ball whose radius is called the \textit{Chebyshev radius}, which we denote $\delta$. We use this to impose some desired regularity on the types of grasps we consider.

\begin{definition}\label{def:valid_wrenches}
    Let $P(\mathcal{X})$ denote the power set of $\mathcal{X}$. The set $\mathcal{S}_\mathcal{O} \subset P(\R^6)$ is the set of all sets of basis wrenches $\mathcal{W}$ with cardinality $n_w$ such that
    \begin{enumerate}
        \item $\mathcal{W}$ is possible to generate on the manipuland $\mathcal{O}$, and
        \item the Chebyshev radii of $\mathcal{C}_w$, denoted $\delta(\mathcal{W})$, are uniformly bounded: $\delta(\mathcal{W})\geq\delta_{min}>0$ for all $\mathcal{W}\in\mathcal{S}_\mathcal{O}$.
    \end{enumerate}
\end{definition}
Condition (2) is mild; for every grasp, $\delta\geq\varepsilon$, so Definition \ref{def:valid_wrenches} excludes grasps that are poor by virtue of a small Chebyshev radius up to a fixed level $\delta_{min}$. This condition is satisfied in practice, since we synthesize grasps by lower bounding their desired quality. This gives the following.

\begin{theorem}\label{thm:min_weight_bound}
    For all basis wrenches $\mathcal{W}\in\mathcal{S}_\mathcal{O}$ on a compact manipuland $\mathcal{O}$, there exists a grasp-independent constant $K(\mathcal{O})>0$ uniformly satisfying $K\ell^*(\mathcal{W}) \leq \varepsilon(\mathcal{W})$.
\end{theorem}

\subsection{Proof of Theorem \ref{thm:min_weight_bound}}
To prove Theorem \ref{thm:min_weight_bound}, we require some intermediate results. Some details are elided to the appendix for brevity. First, consider the following optimization program, whose optimal value is the Ferrari-Canny metric $\varepsilon$.

\begin{lemma}\label{lemma:dai_program_equiv}
    When $0\in\mathcal{C}_w$ (i.e., the grasp is force closure),
    \eqs{\label{eqn:hongkai_program}
        \varepsilon = \minimize_{a \in \R^6, b \in \R}\quad & b \\
        \textnormal{subject to} \quad & a^\top w_l + b \geq 0,\;\forall l \in \mathcal{L}, \\
        &a^\top a = 1.
    }
\end{lemma}
\begin{proof}
    See App. \ref{app:proof_dai_program}.
\end{proof}

We compare the optimal value of \eqref{eqn:hongkai_program} to $\ell^*$ by analyzing the dual of the min-weight LP. By a routine calculation of the dual of \eqref{eqn:min_weight_primal} (see App. \ref{app:min_weight_dual_derivation}) and strong duality, we have
\eqs{\label{eqn:min_weight_dual}
    \ell^* = \minimize_{\nu \in \R^6, \varphi \in \R}\quad & \varphi \\
    \textrm{subject to} \quad & \nu^\top w_l + \varphi \geq 0,\;\forall l \in \mathcal{L} \\
    & \sum_l \parens{\nu^\top w_l + \varphi} = 1.
}

We are now ready to prove Theorem \ref{thm:min_weight_bound}.
\begin{proof}
    We only consider the case where $0\in\mathcal{C}_w$, since otherwise, there is no origin-centered inscribed ball of $\mathcal{C}_w$. This allows us to invoke Lemma \ref{lemma:dai_program_equiv}. Consider an arbitrary set of basis wrenches $\mathcal{W}\in\mathcal{S}_\mathcal{O}$ and suppose $(a^*,b^*)$ is an optimal solution to \eqref{eqn:hongkai_program} (so $\norm{a}_2=1$). The pair $(\nu,\varphi)$ defined
    \eqs{
        \nu = \frac{a^*}{\sum_l \parens{w_l^\top a^* + b^*}}, \quad \varphi = \frac{b^*}{\sum_l \parens{w_l^\top a^* + b^*}}
    }
    is clearly feasible for \eqref{eqn:min_weight_dual}. Dividing by $\sum_l \parens{w_l^\top a^* + b^*}\geq0$ is well-defined, since if the sum were not strictly positive, it would imply that $\mathcal{C}_w$ is degenerate and contained in the hyperplane $\mathcal{P} = \braces{x \mid x^\top a^* + b^* = 0}$, implying $\mathcal{W}\not\in\mathcal{S}_\mathcal{O}$.

    By strong duality and the feasibility\footnote{Since the primal \eqref{eqn:min_weight_primal} is a maximization, the dual \eqref{eqn:min_weight_dual} is a minimization and dual-feasible values \textit{upper bound} the optimal dual and primal values.} of $(\nu,\varphi)$, we have
    \eqs{
        \ell^* = \varphi^* \leq \frac{b^*}{\sum_l \parens{w_l^\top a^* + b^*}} = \frac{\varepsilon}{\sum_l \parens{w_l^\top a^* + b^*}},
    }
    where $b^*=\varepsilon$ by Lemma \ref{lemma:dai_program_equiv}. So, if $K>0$ exists such that
    \eqs{
        K \leq \sum_l \parens{w_l^\top a^*(\mathcal{W}) + b^*(\mathcal{W})}
    }
    uniformly for all $\mathcal{W}\in\mathcal{S}_\mathcal{O}$, the proof follows.
    
    Let $\delta(\mathcal{W})>0$ be the Chebyshev radius of $\mathcal{C}_w$. We show if $v\in\mathcal{C}_w$ is a point in $\mathcal{C}_w$ farthest from $\mathcal{P}$, then $v^\top a^* + b^* \geq 2\delta$. Suppose that $v^\top a^* + b^* < 2\delta$. Let $c$ be a Chebyshev center. Then, the distance of $c$ to $\mathcal{P}$ satisfies $c^\top a^* + b^* \geq \delta$, or $c$ could not be the center of a ball inscribing $\mathcal{C}_w$. Let
    \eqs{
        y := c + \delta a^* \in B_\delta(c) \subset \mathcal{C}_w.
    }
    
    The distance of $y$ to plane $\mathcal{P}$ is given by
    \eqs{
        y^\top a^* + b^* &= c^\top a^* + \delta \cdot (a^*)^\top a^* + b^* \\
        &= \delta + c^\top a^* + b^* \geq 2\delta > a^\top v + b,
    }
    where the last inequality is by assumption, contradicting that $v$ is a farthest point in $\mathcal{C}_w$ from $\mathcal{P}$; thus, $v^\top a^* + b^* \geq 2\delta$.

    Now, there exists an extreme point of $\mathcal{C}_w$ that is a farthest point in $\mathcal{C}_w$ from $\mathcal{P}$, and $\mathcal{W}$ contains all extreme points of $\mathcal{C}_w$ by construction. Thus, there exists $w\in\mathcal{W}$ such that $w^\top a^* + b^* \geq 2\delta$. By Def. \ref{def:valid_wrenches}, we assume a $\delta_\textrm{min}$ such that
    \eqs{
        2\delta_\textrm{min} \leq 2\delta(\mathcal{W}) &\leq \max_{w\in\mathcal{W}} w^\top a^*(\mathcal{W}) + b^*(\mathcal{W}) \\
        &\leq \sum_l \parens{w_l^\top a^*(\mathcal{W}) + b^*(\mathcal{W})},
    }
    so the choice $K=2\delta_{\textrm{min}}$ completes the proof.
\end{proof}

\section{Hardware Experiments}
We now compare real-world grasps synthesized by maximizing the min-weight metric vs. the method of Wu \textit{et al.} \cite{wu2022_learningdexgraspsgenmodel}, which simply finds feasible force closure grasps. Prior work has shown that min-weight synthesized grasps outperform Wu's method under simulated shaking \cite{li2023_frogger}, showing that $\ell^*$ is robust to extrinsic uncertainties. Here, we evaluate \textit{intrinsic} robustness to unknown uncertainties arising from, e.g., poor object reconstruction. This provides some preliminary evidence of our theory's usefulness. We leave analysis and testing of other grasp metrics for future work.

\subsection{Grasp Synthesis with Bilevel Nonlinear Optimization}
Both methods synthesize grasps via nonlinear gradient-based optimization. In \cite{li2023_frogger}, the min-weight-based program is called \textit{FRoGGeR}, and has the following form:
\begin{equation}\label{eqn:frogger}\tag{FRoGGeR}
\begin{split}
    \maximize_q\quad& \ell^*(q) \\
    \subto \quad & q_\text{min} \preceq q \preceq q_\text{max} \\
    &\ell^*(q) \geq k_{\ell} / {n_w} \\
    & FK^i(q) \in \partial\mathcal{O},\; i=1,\dots,n_f \\
    & \text{No (non-finger/object) collision.}
\end{split}
\end{equation}
Wu's method \cite{wu2022_learningdexgraspsgenmodel} instead solves a feasibility program:
\begin{equation}\label{eqn:wu}\tag{Wu}
\begin{split}
    \find\quad&q \\
    \subto \quad & q_\text{min} \preceq q \preceq q_\text{max} \\
    &J(q) = 0 \\
    & FK^i(q) \in \partial\mathcal{O},\; i=1,\dots,n_f \\
    & \text{No (non-finger/object) collision,}
\end{split}
\end{equation}
where $J(q)$ is the value of an optimization program that equals 0 when a grasp is force closure \cite[Eqn. 5]{wu2022_learningdexgraspsgenmodel}. 

Both \eqref{eqn:frogger} and \eqref{eqn:wu} are bilevel optimization programs, since $\ell^*(q)$ and $J(q)$ are themselves optimal values of lower-level programs. However, $\ell^*$ is a measure of a grasp's robustness, while $J$ only indicates whether a grasp is force closure. Thus, \eqref{eqn:frogger} also maximizes a grasp's quality, which we show is vital to real-world performance. The last two constraints are handled using implicit functions by letting the surface $\partial\mathcal{O}$ be the 0-level set of a smooth function $s:\R^3 \rightarrow \R$. For details, see \cite{wu2022_learningdexgraspsgenmodel, li2023_frogger}.

\begin{figure}
    \centering
    \includegraphics[width=\linewidth]{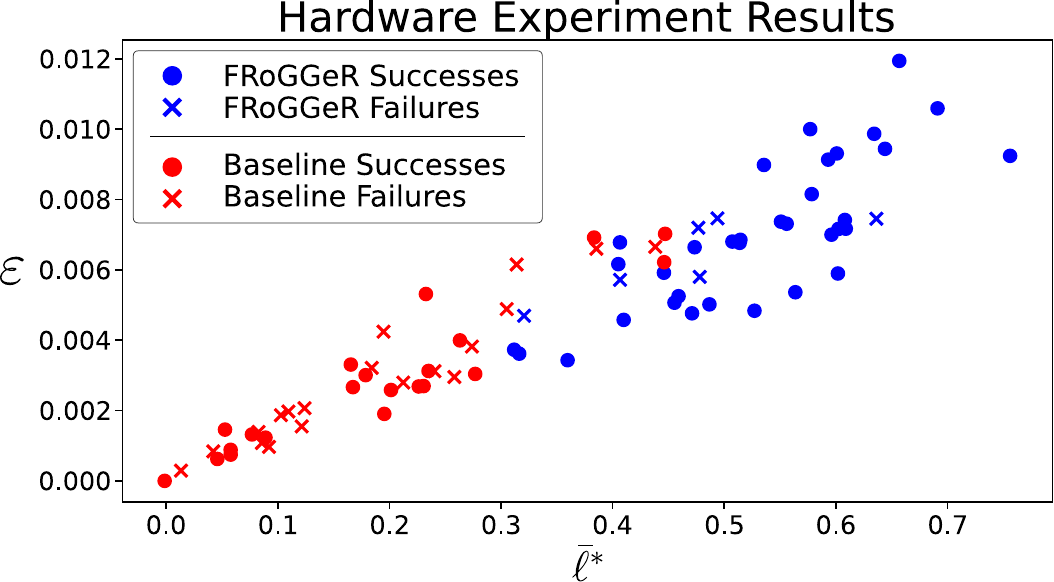}
    \caption{Hardware results. FRoGGeR-synthesized grasps are of higher quality and have a much higher success rate (34/40) than the baseline, which only ensures force closure and does not optimize for grasp quality, resulting in many low-quality grasps and a lower success rate (21/40). This plot also confirms the validity of Theorem \ref{thm:min_weight_bound}, as a linear lower bound can be seen.}
    \vspace{-0.5cm}
    \label{fig:hardware_results}
\end{figure}

\subsection{Experimental Results}
For both methods, we synthesized 5 grasps on 8 objects (see Fig. \ref{fig:hardware_setup}) for a total of 40 grasps each, comparing the success/failure rates as well as the values of \textit{normalized min-weight metric} $\bar{\ell}^*\in[0, 1]$ (where $\bar{\ell}^*:=\ell^*/{n_w}$) and $\varepsilon$. To synthesize grasps using these methods, some object representation is required. We chose to train a NeRF \cite{mildenhall2020_nerf} of the object by scanning the scene with a wrist-mounted camera. Then, a mesh was extracted from the NeRF by running the marching cubes algorithm on a density level set, which we chose empirically to be 17.5 for all objects.

Unlike Wu \textit{et al.}, we did not use a learned model to extract initial guesses for the nonlinear optimization program. Instead, we used the heuristic sampler described in \cite{li2023_frogger} for both methods, and the optimization program we solved was identical except that (a) \eqref{eqn:wu} is a feasibility program, so there is no objective, and (b) \eqref{eqn:wu} enforces a force closure equality constraint, while \eqref{eqn:frogger} enforces that $\bar{\ell}^* \geq k_\ell$, where we used $k_\ell=0.3$ as in \cite{li2023_frogger}.

We used the following simple controller: for an optimal solution $q^*$ of \eqref{eqn:frogger} or \eqref{eqn:wu}, we have corresponding fingertip positions $x^i = FK^i(q^*)$. We defined new target positions for each finger 2\verb|cm| into the surface,
\eqs{
    x_{\text{new}}^i = x^i + 0.02\nabla s(x^i),
}
and let $q^*_{\text{new}}$ satisfy $x_{\text{new}}^i = FK^i(q^*_{\text{new}}),\forall i\in\mathcal{I}$, computed using inverse kinematics and tracked with P control:
\eqs{
    \tau_{\text{fb}} = -K_p(q - q^*_{\text{new}}).
}
After grasping, the object was picked up 20cm and held in place. A grasp was said to fail if it was not successfully picked or if after picking, there was significant rotation.

The quantitative results are summarized in Fig. \ref{fig:hardware_results}. Overall, grasps synthesized with \eqref{eqn:frogger} succeeded in 34/40 trials (85\%), while grasps synthesized using the baseline \eqref{eqn:wu} succeeded in 21/40 (52.5\%). Baseline grasps tended to have low values of $\varepsilon$ and $\bar{\ell}^*$, which supports the theory of uncertainty awareness in grasps detailed in Sec. \ref{sec:theory}. Further, we see that the relation between $\bar{\ell}^*$ and $\varepsilon$ supports Theorem \ref{thm:min_weight_bound}, as a linear lower bound is visually clear.

\subsection{Discussion}
Due to perception error/uncertainty in the NeRF training, the pose and geometry of the object was sometimes inaccurate. This led to some trials where during the pre-grasp motion, the object was struck by the manipulator prior to grasping. These trials were excluded ($\sim$15\% of attempts), since the purpose of the experiments was to evaluate the robustness of grasps themselves. Qualitatively, the dominant failure mode for both methods was attempting grasps on regions where grasp quality was sensitive to shape error. For instance, two failures for FRoGGeR grasps were caused by trying to grab the tea bottle on the sloped area near the top, and many of the failures for Wu's method were caused by trying to grab edges of objects and slipping.

Finally, we note that our reported success rates for Wu's method are significantly lower than than those reported in \cite{wu2022_learningdexgraspsgenmodel}. We suspect that one cause for this is that the initial guesses supplied by Wu's learning-based model are critical for generating good grasps. Additionally, our grasping controller is only position-based, whereas in \cite{wu2022_learningdexgraspsgenmodel}, they command the forces found by computing $J(q)$. The discrepancy could also be due to the different hardware setup, slight differences in implementation, or randomness. Despite these differences, FRoGGeR-synthesized grasps are able to achieve about the same success rate without a learning-based model.

\section{Towards Probabilistic Force Closure}\label{sec:pfc}
The previous analysis applied the framework of intrinsic robustness and Theorem \ref{thm:containment} to characterize the min-weight metric. We now explore another application, and develop the idea of \textit{probabilistic force closure}, wherein we use the relation between object geometry and basis wrenches to apply our theory explicitly when we have a \textit{belief distribution} over the object geometry. We assume we have such a distribution, though obtaining one in practice is challenging and the general case is left for future work.

Suppose the basis wrenches $\mathcal{W}$ associated with a grasp are a random variable generated by, e.g., uncertainty in the contact locations $x^i$, the surface normals $n^i(x^i)$ of $\mathcal{O}$, parameters $\gamma$ of the object $\mathcal{O}(\gamma)$, etc. Then, one can attempt to compute or lower bound the probability of force closure.

Let $\mathcal{\widebar{W}}$ and $\mathcal{W}$ be nominal and random sets of basis wrenches respectively with convex hulls $\mathcal{C}_{\bar{w}}$ and $\mathcal{C}_w$. By Theorem \ref{thm:containment}, we can bound the probability of force closure:
\eqs{\label{eqn:pfc_bound_ineq}
    \mathbb{P}[w_l-\bar{w}_l\in -\mathcal{C}_{\bar{w}}, \; \forall l\in\mathcal{L}] &\leq \mathbb{P}[0 \in \mathcal{C}_w] \\
    &\leq \mathbb{P}[\textrm{force closure}].
}

If we can compute or bound the LHS of \eqref{eqn:pfc_bound_ineq}, then probabilistically-robust grasps may be synthesized. However, as-is, \eqref{eqn:pfc_bound_ineq} is not very useful, since (a) it requires a probability measure to be defined over $\mathcal{O}$, and (b) the relevant uncertainty sets (e.g., $-\mathcal{C}_{\bar{w}}$ in Theorem \ref{thm:containment} or $B_\varepsilon(\bar{w}_l)$ in Corollary \ref{cor:fc_uncertainty_aware}) must be easily integrable with respect to this measure.

\subsection{Probabilistic Object Normals for Grasping}\label{sec:pong}
In light of these challenges, we study a special case where the contact locations $x^i$ are known but the surface normals $n^i$ are random and modeled by a degenerate Gaussian distribution. This gives rise to a method we call \textit{probabilistic object normals for grasping} (PONG).

Let the known contacts be denoted $\posSet=\{\pos^1,\dots,\pos^{n_f}\}$ and consider the set of normals that induce basis wrenches certifying force closure. We call this the \textit{force closure set}:
\eqs{
    \mathfrak{N}(\posSet):=\braces{\{n^i\}_{i=1}^{n_f} \mid 0\in\mathcal{C}_w}. \label{eqn:fc}
}
Suppose the normals are jointly distributed with some density function $\pdf\parens{n^1, \ldots, n^{n_f}}$ and that they are mutually independent such that $\pdf$ can be factorized as $\prod_{i=1}^{n_f} \pdf(n^{i})$. Then, we can compute the probability that $0\in\mathcal{C}_w$:
\eqs{\label{eqn:pfc_intractable}
    \mathbb{P}[0\in\mathcal{C}_w] = \int_{\mathfrak{N}(\posSet)} \prod_{i=1}^{n_f} \pdf(n^i) dn^i. 
}

Though $\pdf$ is factorizable, the integral itself is not, since the integration variables are coupled by the domain $\mathfrak{N}$, and $\mathcal{C}_w$ depends on all of the random normals. Further, $\mathfrak{N}$ is difficult to parameterize; it has no closed form since it is implicitly defined by the condition $0\in\mathcal{C}_w$, where hull membership is checked by solving an LP \cite{li2023_frogger}.

Thus, our strategy is to derive an \textit{approximate force closure set} $\mathcal{A}\subseteq\mathfrak{N}$ in tandem with a choice of density function $p$ for which the integral over $\mathcal{A}$ is known. Since $\mathcal{A}\subseteq\mathfrak{N}$, integrating $\pdf$ over $\mathcal{A}$ yields a lower bound on \eqref{eqn:pfc_intractable}.

\subsection{Random Normals, Forces, and Wrenches}\label{subsec:uncertainty_parameterization}
We model each random normal $n^i$ as the sum of a deterministic \textit{mean normal vector} denoted $\bar{n}^i$ and a \textit{random perturbation vector} $\Delta n^i = \mathcal{T}_i z,$ where $\mathcal{T}_i = \begin{bmatrix}\bar{t}^i_1 & \bar{t}^i_2\end{bmatrix} \in \mathbb{R}^{3 \times 2}$ is a basis for the tangent plane at $\bar{n}^i$, and $z \sim \mathcal{N}(0, \Sigma^i)$ is a zero-mean Gaussian random vector in $\mathbb{R}^2$.

We parameterize the normals in this way for two reasons. First, all perturbations in the direction of $\bar{n}^i$ do not change its direction, and thus do not contribute to uncertainty in its orientation. Second, the planar restriction allows us to integrate a Gaussian density in the plane over a polytope using Green's Theorem \cite[Prop. 1]{hayashi2017_bivariategaussianintegral}.

We remark that with this construction, the random normals $n^i$ will not have unit norm. However, since friction cones are invariant under scaling, as long as the random basis forces $f_j^i$ remain Coulomb-compliant, we can still certify force closure with the induced random basis wrenches.

\begin{figure*}[htbp]
    \centering
    \includegraphics[width=0.49\linewidth]{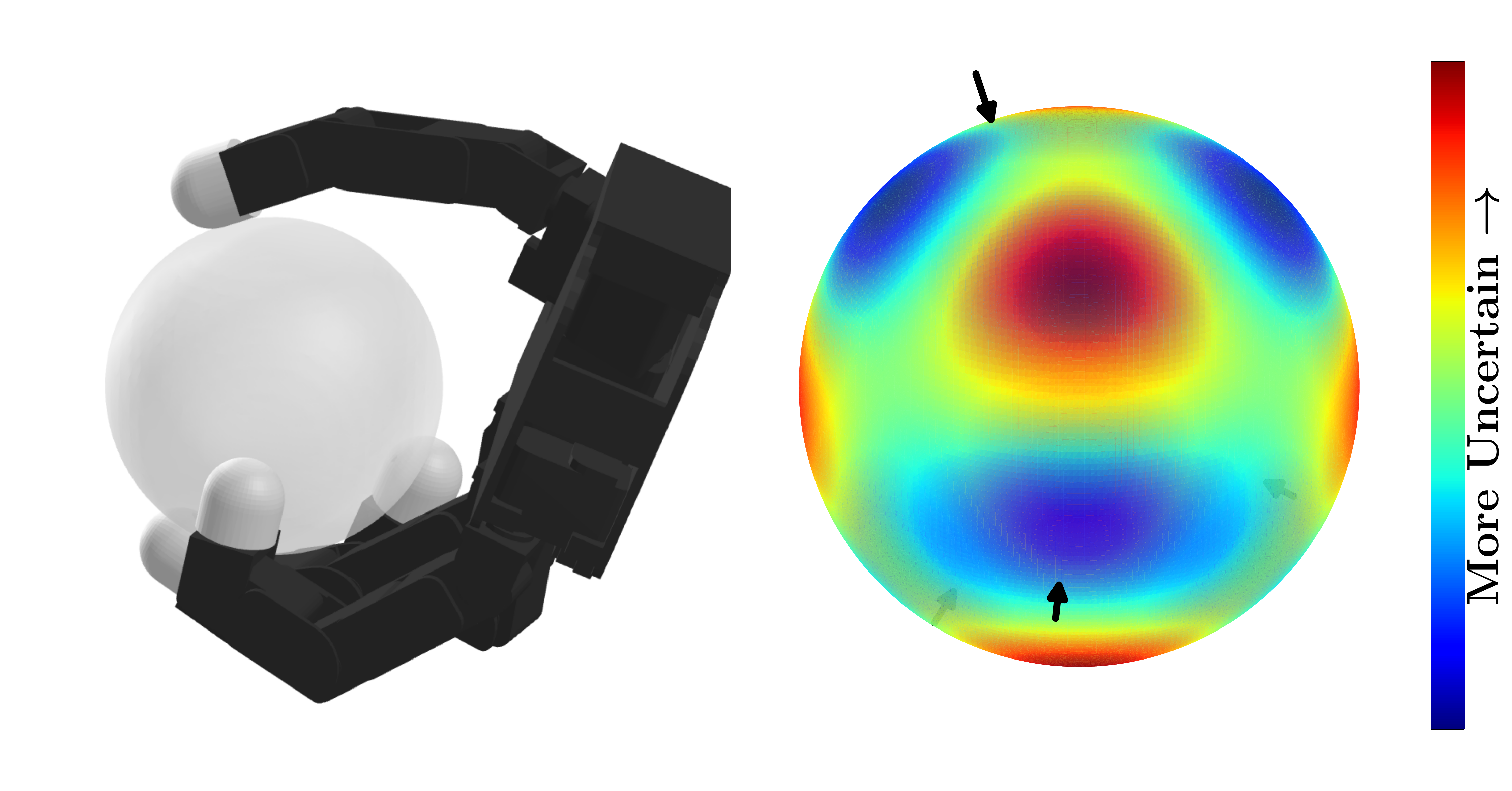}
    \includegraphics[width=0.49\linewidth]{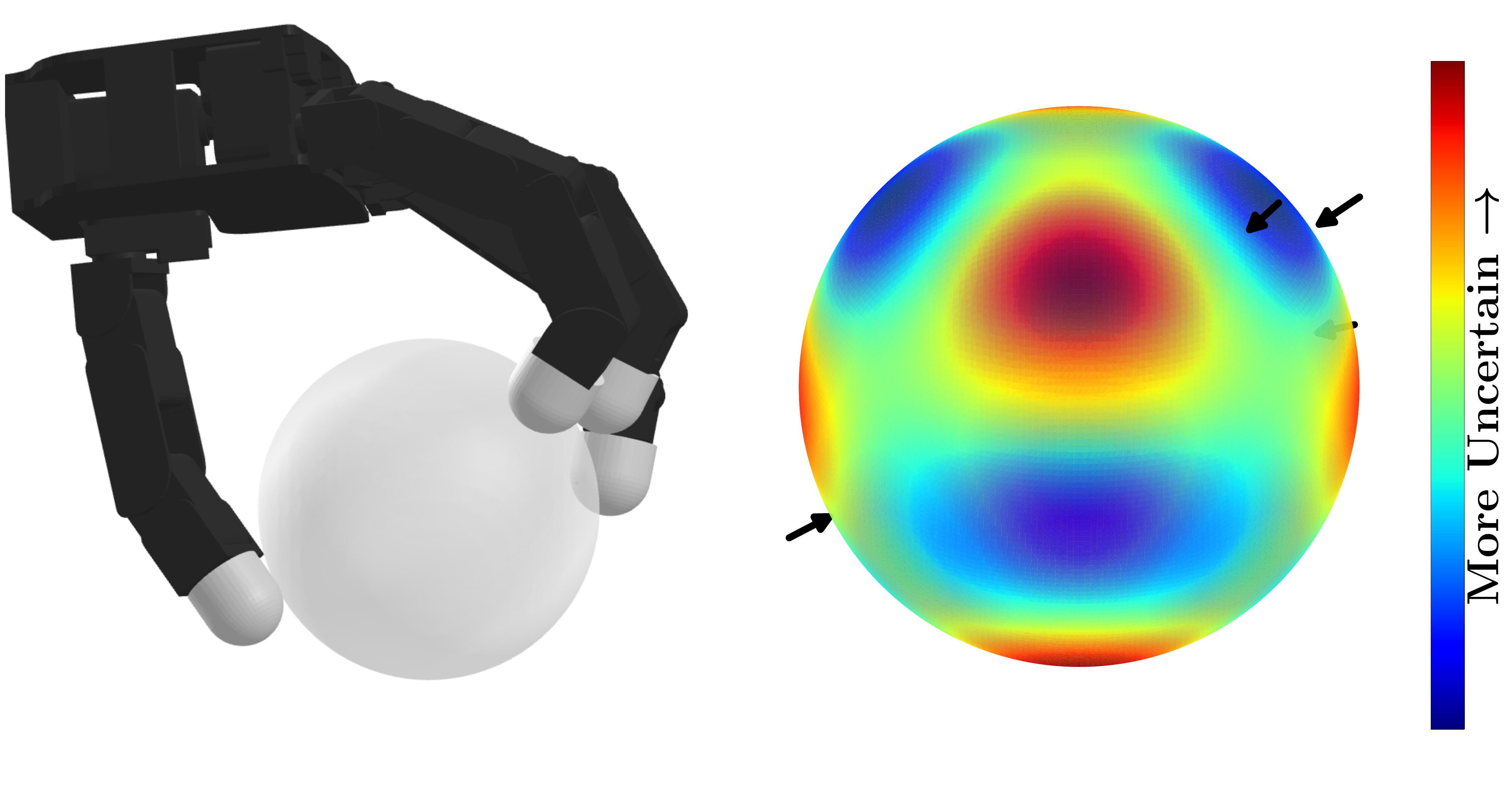}
    \caption{More examples of grasps on a synthetically-uncertain spherical manipuland. The surface normal tangent covariance at each point on the sphere is (proportional to) an isotropic monotonically-increasing function of the real part of the (2,4)-spherical harmonic. The synthesized grasps tend to avoid placing the fingertips on the red uncertain regions while remaining kinematically feasible. They were synthesized in 6.47 and 4.80 seconds respectively.}
    \label{fig:sph_harm_grasps}
    \vspace{-0.5cm}
\end{figure*}

We now construct a random friction pyramid about the random normal $n^i$ such that its edges are always Coulomb-compliant. First, consider the \textit{mean} basis wrenches $\bar{w}_j^i$ with force and torque components $\bar{f}_j^i$ and $\bar{\tau}_j^i$. We represent $\bar{f}_j^i$ as the sum of $\bar{n}^i$ and a tangent component $\parens{\bar{f}_j^i}_t$ of length $\mu$. We compute $\parens{\bar{f}_j^i}_t$ using a unit length \textit{generator} $\bar{g}_j^i(\bar{n}^i)\in\R^3$ of our choice orthogonal to $\bar{n}^i$ such that
\eqs{
    \parens{\bar{f}_j^i}_t = \mu\cdot\parens{\bar{g}_j^i \times \bar{n}^i}.
}
For example, for some arbitrary $v\neq\bar{n}^i$, we could pick $\bar{g}_j^i(\bar{n}^i)=(v \times \bar{n}^i)/\norm{v \times \bar{n}^i}_2$. Thus, we can write
\eqs{
    \bar{f}_j^i &= \bar{n}^i + \mu\cdot\parens{\bar{g}_j^i \times \bar{n}^i} \\
    \implies \bar{w}_j^i &= \underbrace{\mat{\parens{I + \mu\cdot\wedgeop{\bar{g}_j^i}} \\ \wedgeop{x^i}\parens{I + \mu\cdot\wedgeop{\bar{g}_j^i}}}}_{:=T_j^i(\bar{n}^i)}\bar{n}^i,
}
and similarly, $w_j^i = T_j^in^i$, which satisfies $\norm{(f_j^i)_t}_2\leq\mu\norm{n^i}$ $\big($in contrast, recall that $\norm{\parens{\bar{f}_j^i}_t}_2=\mu\norm{n^i}$\big). To see this, let $\psi_j^i$ be the angle between $\bar{g}_j^i$ and $n^i$. Then,
\eqs{
    \norm{(f_j^i)_t}_2 = \mu\norm{\bar{g}_j^i}\norm{n^i}\sin(\psi_j^i)\leq\mu\norm{n^i},
}
ensuring Coulomb-compliance of the random basis forces.

\subsection{Deriving an Approximate Force Closure Set}\label{subsec:decoupling}
Next, we define disjoint sets $\mathcal{A}^i$ whose union defines a set $\mathcal{A}\subseteq\mathfrak{N}$. Combining Theorem \ref{thm:containment} with the definition of $\mathfrak{N}$ and the relation $w^i_j = T^i_jn^i$ from Sec. \ref{subsec:uncertainty_parameterization}, if $\mathcal{A}^i$ satisfies
\eqs{\label{eqn:decomposed_fc_sets}
    \mathcal{A}^i \subseteq \{ n^i \mid T^i_j (n^i - \bar{n}^i) \in \mathcal-{C}_{\bar{w}},\;\forall j\in\mathcal{J}\},
}
then we have
\eqs{
    n^i\in\mathcal{A}^i,\;\forall i\in\mathcal{I} \implies (n^1,\ldots,n^{n_f})\in\mathfrak{N}.
}
Letting $\mathcal{A}=\bigcup_{i=1}^{n_f}\mathcal{A}^i$, we therefore have the bound
\eqs{\label{eqn:decomposed_integral}
    \prod_{i=1}^{n_f}\int_{\mathcal{A}^i}p(n^i)dn^i &= \int_{\mathcal{A}}\prod_{i=1}^{n_f}p(n^i)dn^i \\
    &\leq \int_{\mathfrak{N}}\prod_{i=1}^{n_f}p(n^i)dn^i.
}

We can factorize the integral this way since $\mathcal{C}_{\bar{w}}$ in \eqref{eqn:decomposed_fc_sets} depends only on the \textit{mean} normals $(\bar{n}^1,\ldots,\bar{n}^{n_f})$, which are fixed parameters of the known uncertainty distributions. In contrast, in \eqref{eqn:fc}, $\mathcal{C}_w$ (and thus $\mathfrak{N}$) depends jointly on all of the \textit{random} normals, which are the integration variables.

Unfortunately, the set on the RHS of \eqref{eqn:decomposed_fc_sets} still cannot be expressed in closed form. Thus, we let $\mathcal{A}^i$ conservatively approximate it by computing a polytope with large volume satisfying \eqref{eqn:decomposed_fc_sets} that is parameterized by its vertices.

To do this, for each finger, we fix $k$ search directions $d^{i,k}\in\R^3$ in the tangent plane to $\mathcal{O}$ at $x^i$ with index set $\mathcal{K}$. We compute the longest step $\theta^{i,k}\geq0$ that can be taken in this direction while satisfying \eqref{eqn:decomposed_fc_sets} and denote the point $\theta^{i,k} d^{i,k}$ as $v^{i,k}$. We then let $\mathcal{A}^i$ be the convex hull over all $v^{i,k}$.

Conveniently, this search can be expressed as a linear program. Define $\widebar{W}\in\R^{6\times n_w}$, whose columns are the nominal basis wrenches. Then, to compute each $v^{i,k}$, we can solve the following LP for each $(i,k)\in\mathcal{I} \times \mathcal{K}$ in parallel:
\begin{subequations}
\label{opt:vlp}
\begin{align}
    \maximize_{\substack{\theta^{i,k} \in \R \\ \braces{\alpha^{i,k}_j}_{j=1}^{n_s} \subset \R^{n_w}}} \quad & \theta^{i,k} \\
    \subto \quad & \theta^{i,k} \geq 0 \label{constr:vlp_delta} \\
    & \alpha^{i,k}_j \succeq 0,\;\forall j\in\mathcal{J} \label{constr:vlp_alpha1} \\
    & \mathds{1}^{\top}_{n_w} \alpha^{i,k}_j = 1,\;\forall j\in\mathcal{J} \label{constr:vlp_alpha2} \\
    & T_j^i\parens{\theta^{i,k} d^{i,k}} = -\widebar{W}\alpha^{i,k}_j,\;\forall j\in\mathcal{J}. \label{constr:cvx}
\end{align}
\end{subequations}
Constraint \eqref{constr:vlp_delta} enforces nonnegativity of the scaling along $d^{i,k}$, constraints \eqref{constr:vlp_alpha1} and \eqref{constr:vlp_alpha2} enforce that the $\alpha^{i,k}_j$ are valid convex weights, and constraint \eqref{constr:cvx} enforces that the random normal $n^i = \bar{n}^i + \theta^{i,k} d^{i,k}$ must lie in $\mathcal{A}^i$. 

Therefore, after solving \eqref{opt:vlp}, we can compute
\eqs{\label{eqn:final_factorized_bound}
    L_{FC} := \prod_{i=1}^{n_f}\int_{\mathcal{A}^i(\theta,d)}p(n^i)dn^i
}
using \cite[Prop. 1]{hayashi2017_bivariategaussianintegral}, which bounds the probability of force closure. Due to space constraints, computational considerations for computing \eqref{eqn:final_factorized_bound} are detailed in the appendix.

\subsection{Toy Examples}
We first demonstrate PONG on two toy examples. In both, we let the manipuland $\mathcal{O}$ be a sphere of radius $0.05$\verb|m|, and we define (unnormalized) distributions over $x\in\partial\mathcal{O}$ as follows:
\eqs{
    &\bar{n}(x) = x / \norm{x}_2, \\
    &\sigma_1^2 = 100x_3^2,\quad \sigma_2^2 = 0.01\exp\parens{Y^2_4(\theta(x), \phi(x))},
}
where the mean of both is $\bar{n}(x)$ and they have isotropic tangent variances defined by $\sigma_1^2$ and $\sigma_2^2$ respectively.

The first distribution assigns low uncertainty at the equator of the object, symmetrically increasing it towards the poles. The second assigns uncertainties based on the (2,4)-spherical harmonic $Y^2_4$, where $\theta(x),\phi(x)$ are the azimuthal and polar angles of $x\in\partial\mathcal{O}$. Visualizations of grasps generated with each are shown in Fig. \ref{fig:banner} and Fig. \ref{fig:sph_harm_grasps} respectively. 

We note the qualitative difference that the uncertainty distribution makes, even though the manipuland's geometry is identical in each case. This demonstrates that accounting for uncertainty in grasp synthesis can yield vastly different solutions: for example, a vision system could report higher uncertainty for regions of the object that are occluded, which in turn could encourage grasping visible regions.

\subsection{Comparative Analysis on Realistic Meshes}\label{sec:realistic_sim_results}
While PONG-synthesized grasps on toy examples showed some promise, to use it on more realistic examples, we need a distribution supported on $\partial\mathcal{O}$. In the absence of a natural choice, we define a \textit{curvature-regularized} one. At $x\in\partial\mathcal{O}$, the \textit{shape operator} $\mathfrak{S}_x:T_x\partial\mathcal{O} \rightarrow T_x\partial\mathcal{O}$ is defined as the directional derivative $\mathfrak{S}_x(v):=-D_vN(x)$, where $N$ is the unit normal at $x$. The eigenpairs of $\mathfrak{S}_x$, $(\kappa_1,v_1),(\kappa_2,v_2)\in \R \times T_x\partial\mathcal{O}$, are the \textit{principal curvatures and directions} at $x$ respectively \cite{oneill2006_diffgeom}.

Since $N(x)=\nabla s(x)/\norm{\nabla s(x)}_2$, we have
\eqs{
    \mathfrak{S}_x(v) = -\parens{I - \frac{\nabla s(x)\nabla s(x)^\top}{\norm{\nabla s(x)}_2^2}}\frac{\nabla^2 s(x)}{\norm{\nabla s(x)}_2}v.
}
Letting the directions $v_1,v_2$ form our tangent basis at each contact $x^i$ and (some increasing function of) the magnitude of $\kappa_1,\kappa_2$ form our variances, we can define a curvature-sensitive uncertainty distribution with parameters
\eqs{\label{eqn:curvature_distribution}
    \bar{n}^i &:= - \brackets{\nabla_x s(x^i) / \norm{\nabla_x s(x^i)}_2}, \\
    \bar{t}^i_m &:= v_m,\; m\in\{1,2\}, \\
    \parens{\sigma^i_m}^2 &:= \log\parens{K_{\text{curv}}\cdot\abs{\kappa_{m}} + h},\;m\in\{1,2\},
}
where $K_{\text{curv}}>0$ relates curvature to uncertainty and $h>0$ captures prior uncertainty at $x$. Intuitively, we assign greater uncertainty to high curvatures, since slight errors in position will lead to large changes in the surface normal.

Since $\ell^*$ captures intrinsic uncertainty, we study whether it performs comparably to PONG even though it only provides conservative guarantees on uncertainty-awareness per Corollary \ref{cor:fc_uncertainty_aware} without reasoning about the bound \eqref{eqn:pfc_bound_ineq}, and find that it does. We synthesize 2000 grasps on 5 objects (a camera, teacup, rotated Rubik's cube, tape dispenser, and teddy bear) and in simulation study pick success as a function of metric.

\begin{figure}
    \centering
    \includegraphics[width=\linewidth]{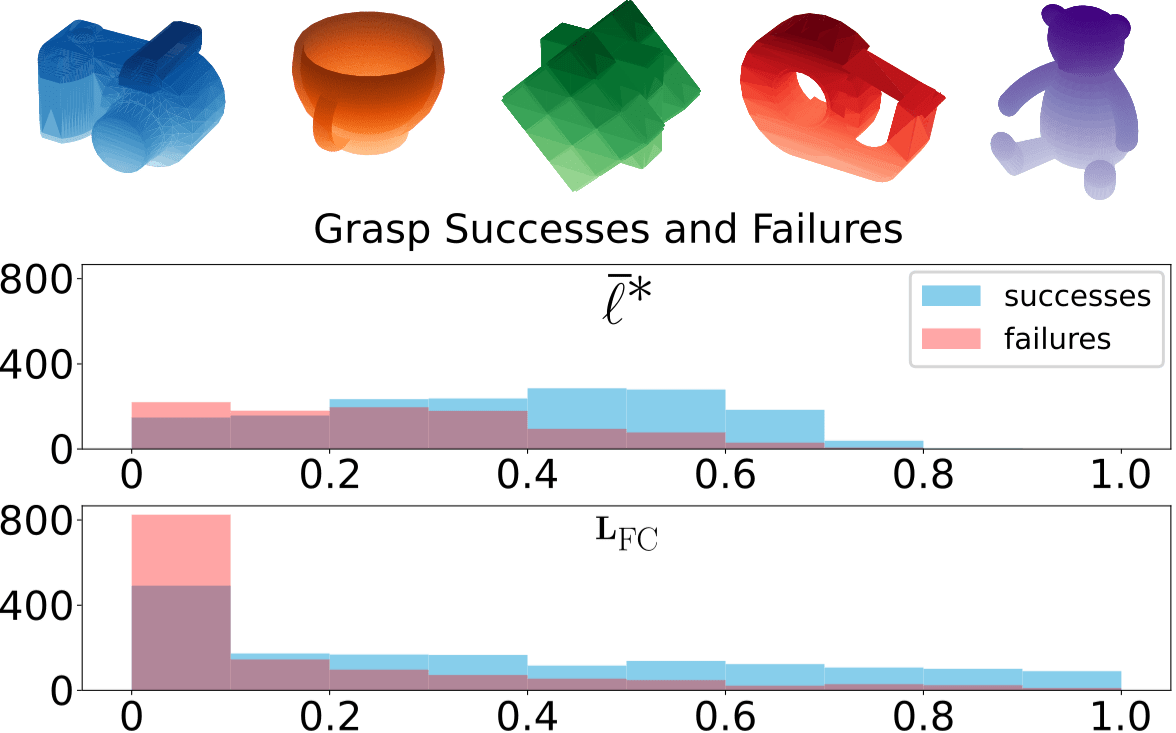}
    \caption{\textbf{Top.} 5 simulated objects. \textbf{Bottom.} Success/failure histograms on 2000 simulated grasps for $\ell^*$ and $L_{FC}$.}
    \label{fig:success_failure}
    \vspace{-0.5cm}
\end{figure}

All 2000 grasps are synthesized using \eqref{eqn:frogger} but with objective $L_{FC}(q)$ and evaluated on $\ell^*$ and $L_{FC}$. We observe two key trends (see Fig. \ref{fig:success_failure}). Based on the theory of Sec. \ref{sec:theory}, we expect fewer failures as each metric rises, which we observe. Further, because each metric is conservative, we expect the ratio of successes to failures to rise with the metric, which is again confirmed. This provides evidence that the min-weight metric is empirically probabilistically sensitive to certain uncertainties in object geometry.

\section{Conclusion}
This work develops a theory of intrinsic uncertainty and robustness for dexterous grasps. Theorems \ref{thm:containment} and \ref{thm:min_weight_bound} provide guarantees on the intrinsic robustness of $\ell^*$-synthesized grasps, motivating its usage as a principled and efficient metric. These guarantees were tested on hardware against a competitive baseline, demonstrating the importance of robustness in grasp synthesis. We also use our theory to develop PONG, a method for probabilistically-aware grasp synthesis. We show that PONG can generate grasps that account for uncertainty over the object surface, paving the way for using stochastic models of geometric uncertainty with grasp planners. In future work, we will study acquiring uncertainty distributions compatible with PONG to more explicitly account for intrinsic uncertainty in dexterous grasps.

\bibliographystyle{unsrt}
\bibliography{references}

\newpage

\appendix

\subsection{Proof of Lemma \ref{lemma:dai_program_equiv}}\label{app:proof_dai_program}
Program \eqref{eqn:hongkai_program} is motivated by the following result, which is rephrased from \cite[Eqn. (27)]{dai2015_forceclosuresdp} using this paper's notation.
\begin{lemma}\label{lemma:dai_condition}
    Fix some $r \geq 0$ and points $\mathcal{W} = \{w_l\}_{l=1}^{n_w}\subset\mathbb{R}^6$. Then, $B_r(0) \subset \mathcal{C}_w$ if and only if $b\geq r$ for every pair $(a,b)\in\mathbb{R}^6\times\mathbb{R}$ that satisfies
    \eqs{
        &a^\top w_l + b \geq 0,\;\forall l \in \mathcal{L}, \\
        &a^\top a = 1.
    }
\end{lemma}
The choice $r=\varepsilon$ proves Lemma \ref{lemma:dai_program_equiv} by contradiction.
\begin{proof}[Proof of Lemma \ref{lemma:dai_program_equiv}]
    Let $(a^*,b^*)$ be an optimal solution for \eqref{eqn:hongkai_program} and let $\mathcal{H}:=\{x \mid x^\top a^* + b^* \geq 0\}$.
    
    First, we show $b^*\geq0$. Suppose that $b^*<0$. Then, $0\not\in\mathcal{H}$, since $0^\top a^* + b^*=b^* < 0$. However, for $(a^*,b^*)$ to be feasible, we must have that $\mathcal{C}_w \subset \mathcal{H}$ by the first constraint of \eqref{eqn:hongkai_program}. This contradicts the assumption that $0\in\mathcal{C}_w$, so $b^* \geq 0$.
    
    Now, suppose for the sake of contradiction that $b^*\neq\varepsilon$. Since $(a^*,b^*)$ satisfies Lemma \ref{lemma:dai_condition}, $B_{b^*}(0)\subset\mathcal{C}_w$ and $\norm{a}_2=1$. Further, by definition, $B_\varepsilon(0)\subset\mathcal{C}_w$. Consider two cases.
    \begin{itemize}
        \item \textbf{\underline{Case: $b^* < \varepsilon$.}} Because $b^*<\varepsilon$ and $B_\varepsilon(0)\subset\mathcal{C}_w$, we must have that $-b^*a^*\in B_{b^*}(0)\subset\setint(\mathcal{C}_w)$ and $-\varepsilon a^*\in B_{\varepsilon}(0)\subset\mathcal{C}_w$. By convexity, the point
        \eqs{
            y = -\frac{(b^*+\varepsilon)}{2}a^* 
        }
        lies in $\mathcal{C}_w$. First, we must have $y\not\in\mathcal{H}$, since
        \eqs{\label{eqn:contradiction_eqn_1}
            y^\top a^* + b^* &= \frac{1}{2}(b^*-\varepsilon) < 0,
        }
        where $b^*-\varepsilon<0$ by assumption. However, $y\in\mathcal{C}_w$, so there must exist convex weights $\alpha$ such that $y=\sum_l\alpha_l w_l$. This implies that $y\in\mathcal{H}$, because
        \eqs{\label{eqn:contradiction_eqn_2}
            y^\top a^* + b^* &= \sum_l\alpha_l w_l^\top a^* + \sum_l\alpha_l b^* \\
            &= \sum_l\alpha_l\parens{w_l^\top a^* + b^*} \geq 0
        }
        since $(a^*,b^*)$ is a feasible solution to \eqref{eqn:hongkai_program}. Equations \eqref{eqn:contradiction_eqn_1} and \eqref{eqn:contradiction_eqn_2} contradict, so we cannot have $b^*<\varepsilon$.

        \item \textbf{\underline{Case: $b^* > \varepsilon$.}} Since $\varepsilon$ is the radius of the largest ball inscribed in $\mathcal{C}_w$, there exists a facet of $\mathcal{C}_w$ with outward pointing unit normal $d^*$ such that $\varepsilon d^*\in\partial\mathcal{C}_w$, which defines a supporting hyperplane $\{x \mid -x^\top d^* + \varepsilon = 0\}$ of $\mathcal{C}_w$. Therefore, $(-d^*,\varepsilon)$ is a feasible solution for \eqref{eqn:hongkai_program}, since $\mathcal{C}_w$ is the intersection of all halfspaces containing $\mathcal{W}$, and every facet of $\mathcal{C}_w$ is generated by a supporting hyperplane of $\mathcal{C}_w$. However, the case $\varepsilon<b^*$ contradicts that $(a^*,b^*)$ is optimal for \eqref{eqn:hongkai_program}, completing the proof.
    \end{itemize}
\end{proof}

\subsection{A Detailed Derivation of the Min-Weight LP Dual}\label{app:min_weight_dual_derivation}
Recall the definition of the min-weight metric:
\eqs{
    \ell^* = \maximize_{\alpha,\ell}\quad & \ell \\
    \subto \quad & W\alpha = 0 \\
    & \mathds{1}^\top \alpha = 1 \\
    & \alpha \succeq \ell\mathds{1}.
}
We will derive the dual. First, note that the min-weight metric can be rewritten as
\eqs{
    \ell^* = -\minimize_{\alpha,\ell}\quad & -\ell \\
    \subto \quad & W\alpha = 0 \\
    & \mathds{1}^\top \alpha = 1 \\
    & \alpha \succeq \ell\mathds{1}.
}
The Lagrangian of the minimization can be written as
\eqs{
    \mathfrak{L} &= -\ell + \lambda^\top(\ell\mathds{1} - \alpha) + \nu^\top W\alpha + \varphi(\mathds{1}^\top \alpha - 1) \\
    &= \ell(-1 + \lambda^\top\mathds{1}) + \alpha^\top(-\lambda + W^\top\nu + \varphi\mathds{1}) - \varphi.
}
The dual function is therefore
\eqs{
    g(\lambda,\nu,\varphi) = \begin{cases}
        -\varphi, & \textrm{if } \lambda^\top\mathds{1} = 1, \; W^\top\nu + \varphi\mathds{1} = \lambda, \\
        -\infty, & \textrm{else}.
    \end{cases}
}
Thus, we have the dual program, and by strong duality,
\eqs{
    \ell^* = -\maximize_{\lambda,\nu,\varphi}\quad & -\varphi \\
    \subto \quad & \mathds{1}^\top\lambda = 1 \\
    & \lambda = W^\top\nu + \varphi\mathds{1} \\
    & \lambda \succeq 0.
}
Rewriting the dual as a minimization and eliminating $\lambda$ yields the program \eqref{eqn:min_weight_dual}:
\eqsnn{
    \ell^* = \minimize_{\nu,\varphi}\quad & \varphi \\
    \subto \quad & \nu^\top w_l + \varphi \geq 0,\;\forall l \in \mathcal{L} \\
    & \sum_l \parens{\nu^\top w_l + \varphi} = 1.
}

\subsection{Integrating Planar Gaussians Over Polytopes}\label{app:bigauss_proof}
Sec. \ref{sec:pong} references the ability to integrate a Gaussian probability density over a polytope. This result makes the procedure precise.
\begin{proposition}\label{prop:bigauss}
    Let $\mathscr{P}$ be a planar polygon with $n_v$ vertices $y^1,\dots,y^{n_v}\in\R^2$ ordered counterclockwise with $y^{{n_v}+1}:=y^1$. Let $Z$ be a bivariate Gaussian random variable with mean $\mu$ and covariance $\Sigma=\diag(\sigma_1^2,\sigma_2^2)$. Then, 
    \eqs{\label{eqn:integral}
        \mathbb{P}[Z\in\mathscr{P}] = \frac{1}{\sigma_2\sqrt{8\pi}}\sum_{m=1}^{n_v} D^m\int_0^1A^m(r)B^m(r)dr,
    }
    where the terms above are
    \eqsnn{
        D^m &:= y_2^{m+1} - y_2^m, \\
        A^m(r) &:= \exp\parens{-\frac{1}{2\sigma_2^2}\brackets{(1-r)y_2^m + ry_2^{m+1} - \mu_2}^2}, \\
        B^m(r) &:= \erf\parens{\frac{(1-r)y_1^m+ry_1^{m+1} - \mu_1}{\sigma_1\sqrt{2}}}.
    }
\end{proposition}
The proof of Proposition \ref{prop:bigauss} closely follows the one presented in \cite[Proposition 1]{hayashi2017_bivariategaussianintegral}. First, we recall Green's Theorem.
\begin{theorem}[Green's Theorem]\label{thm:greens_thm}
    Let $\mathcal{D}$ be a closed region in the plane with piecewise smooth boundary. Let $P(y_1,y_2)$ and $Q(y_1,y_2)$ be continuously differentiable functions defined on an open set containing $\mathcal{D}$. Then,
    \eqs{
        &\oint_{\partial\mathcal{D}}P(y_1,y_2)dy_1 + Q(y_1,y_2)dy_2 \\
        &\quad = \iint_{\mathcal{D}}\parens{\frac{\partial Q}{\partial y_1} - \frac{\partial P}{\partial y_2}}dy_1dy_2.
    }
\end{theorem}
Second, we prove a useful intermediate result.
\begin{lemma}\label{lemma:exp_integral}
The following holds:
    \eqs{\label{eqn:lemma_integral_eqn}
        &\int\exp(-(ay^2+2by+c))dy \\
        &\quad = \frac{1}{2}\sqrt{\frac{\pi}{a}}\exp\parens{\frac{b^2-ac}{a}}\erf\parens{\sqrt{a}y + \frac{b}{\sqrt{a}}} + \textrm{const.}
    }
\end{lemma}
\begin{proof}
    Recall that
    \eqs{
        \erf(y) = \frac{2}{\sqrt{\pi}}\int_0^y\exp(-t^2)dt.
    }
    We differentiate the RHS of \eqref{eqn:lemma_integral_eqn} with respect to $y$ and by the Fundamental Theorem of Calculus, we have
    \eqs{
        &\frac{1}{\sqrt{a}}\exp\parens{\frac{b^2-ac}{a}}\cdot\sqrt{a}\exp\parens{-\parens{\sqrt{a}y + \frac{b}{\sqrt{a}}}^2} \\
        &\quad= \exp\parens{\frac{b^2-ac}{a}}\exp\parens{-\parens{ay^2+2by+\frac{b^2}{a}}} \\
        &\quad= \exp(-(ay^2+2by+c)),
    }
    which is the integrand of the LHS, proving the claim.
\end{proof}
We are now ready to prove Proposition \ref{prop:bigauss}.
\begin{proof}[Proof of Proposition \ref{prop:bigauss}]
    We have that
    \eqs{
        &(y-\mu)^\top\Sigma^{-1}(y-\mu) \\
        &\quad= \frac{1}{\sigma_1^2}y_1^2 - \frac{2}{\sigma_1^2}\mu_1y_1 + \brackets{\frac{1}{\sigma_2^2}(y_2-\mu_2)^2 + \frac{\mu_1^2}{\sigma_1^2}}.
    }
    Letting
    \eqs{
        a &= \frac{1}{2\sigma_1^2},\\
        b &= \frac{-\mu_1}{2\sigma_1^2},\\
        c &= \frac{1}{2}\brackets{\frac{1}{\sigma_2^2}(y_2-\mu_2)^2 + \frac{\mu_1^2}{\sigma_1^2}},
    }
    and applying Lemma \ref{lemma:exp_integral} to the bivariate Gaussian density function $f(y_1,y_2)$, we have
    \eqs{
        &\int f(y_1,y_2)dy_1 \\
        &= \frac{1}{2\sigma_2\sqrt{2\pi}}\exp\parens{-\frac{1}{2}\parens{\frac{y_2 - \mu_2}{\sigma_2}}^2}\erf\parens{\frac{y_1-\mu_1}{\sigma_1\sqrt{2}}} \\
        &\quad+ \textrm{const}.
    }
    Applying Theorem \ref{thm:greens_thm}, we see that for the choice $P=0$ and $Q=\int f(y_1,y_2)dy_1$,
    \eqs{
        &\iint_\mathcal{D}f(y_1,y_2)dy_1dy_2 \\
        &= \oint_{\partial\mathcal{D}}\frac{1}{2\sigma_2\sqrt{2\pi}}\exp\parens{-\frac{\parens{y_2 - \mu_2}^2}{2\sigma_2^2}}\erf\parens{\frac{y_1-\mu_1}{\sigma_1\sqrt{2}}}dy_2.
    }
    To evaluate the contour integral, we split the contour up into the line segments formed by connecting the $n_v$ extreme points of $\mathcal{D}$ in counterclockwise order. A point $y$ on the $m^{th}$ segment can be expressed
    \eqs{
        y = (1-r)\mat{y_1^m \\ y_2^m} + r\mat{y_1^{m+1} \\ y_2^{m+1}},\; r\in[0,1].
    }
    Performing this change of variables and letting $y^{n_v+1}=y^1$,
    \eqs{
        &\iint_\mathcal{D}f(y_1,y_2)dy_1dy_2 \\
        &\quad= \frac{1}{\sigma_2\sqrt{8\pi}}\sum_{m=1}^{n_v} D^m\int_0^1A^m(r)B^m(r)dr,
    }
    where
    \eqsnn{
        D^m &:= y_2^{m+1} - y_2^m, \\
        A^m(r) &:= \exp\parens{-\frac{1}{2\sigma_2^2}\brackets{(1-r)y_2^m + ry_2^{m+1} - \mu_2}^2}, \\
        B^m(r) &:= \erf\parens{\frac{(1-r)y_1^m+ry_1^{m+1} - \mu_1}{\sigma_1\sqrt{2}}}.
    }
    Finally, noting that $\mathbb{P}[Z\in\mathcal{D}]=\iint_\mathcal{D}f(y_1,y_2)dy_1dy_2$ completes the proof.
\end{proof}

\subsection{Efficiently Solving Batches of Vertex LPs}
Program \eqref{opt:vlp} suggests solving a batch of $n_f \cdot n_v$ linear programs in parallel to compute the scaling values $\theta^{i,k}\in\R$. In fact, it is equivalent to further parallelize the LP computation in the following way. 

\begin{proposition}[Efficient Vertex LP Batching]
    The following equality holds:
    \eqs{
        \theta^{i,k} = \min_{j=1,\dots,n_s}\theta^{i,k}_j,
    }
    where (with a slight abuse of notation) $\theta^{i,k}_j$ is the optimal solution to the following LP for a fixed index triple $(i,j,k)$.
    \begin{subequations}
    \label{opt:vlp_efficient}
    \begin{align}
        \maximize_{\theta^{i,k}_j \in \R, \; \alpha^{i,k}_j \in\R^{n_w}} \quad& \theta^{i,k}_j \\
        \subto \quad & \theta^{i,k} \geq 0 \\
        & \alpha^{i,k}_j \succeq 0 \\
        & \mathds{1}^\top \alpha^{i,k}_j = 1 \\
        & \parens{\theta^{i,k} d^{i,k}} T_j^i = -\widebar{W}\alpha^{i,k}_j.
    \end{align}
    \end{subequations}
\end{proposition}

To actually solve many LPs in a batched manner, we use a custom port of the \verb|quantecon| implementation of the simplex method, available at the following link: \href{https://github.com/alberthli/jax_simplex}{github.com/alberthli/jax\_simplex}. Because this implementation is in \verb|JAX|, it can be run on both CPU or GPU without any additional modification. Due to certain parts of our computation stack being CPU-bound, we choose to compute the bound entirely on CPU. We observed that an \verb|Intel i9-12900KS| CPU typically exhibited about a 20\% increase in speed over an \verb|AMD Ryzen Threadripper PRO 5995WX|, which we attribute to speedups in Intel vs. ARM architectures on BLAS routines.

\subsection{\texorpdfstring{Differentiating $L_{FC}$}{Differentiating LFC}}
In order to maximize the bound in \eqref{eqn:final_factorized_bound} in a gradient-based nonlinear optimization program, we must compute the gradient of $L_{FC}$ with respect to the robot configuration $q$. To accomplish this, we need three major components:
\begin{itemize}
    \item differentiating through the numerical integration scheme used to evaluate the expressions in Proposition \ref{prop:bigauss} with respect to the polygon vertices $v^{i,k}$;
    \item differentiating through VLP$^{i,k}$ in \eqref{opt:vlp} (or the more efficient program \eqref{opt:vlp_efficient}) with respect to the wrench matrix $W(q)$;
    \item differentiating through the parameters $W$, $\bar{n}^i$, $\{\bar{t}^i_1,\bar{t}^i_2\}$, and $\{\sigma^i_1, \sigma^i_1\}$ with respect to the configuration $q$.
\end{itemize}

To differentiate through the numerical integration, we use the open source package \verb|torchquad| designed to differentiate numerical quadrature methods. All sub-expressions in Proposition \ref{prop:bigauss} can be implemented in \verb|JAX|, which yields the desired gradient in a straightforward manner.

To differentiate the optimal value of $\delta^{i,k}$ in \eqref{opt:vlp_efficient} with respect to the robot configuration $q$, we use implicit differentiation of the KKT conditions. The analytical gradient can be computed quickly in a nearly identical way to the one used by FRoGGeR, since here we also solve a linear program \cite[Prop. 1]{li2023_frogger}. For the the case of quadratic programs and more general programs, see \citeapp{amos2017_diffopt}.

We compute the gradients with respect to the uncertainty distribution parameters directly using \verb|JAX|. However, due to the special structure of \eqref{opt:vlp_efficient}, we compute the gradients of $W$ with respect to the distribution parameters completely analytically, which in practice leads to a large speedup. Because deriving these gradients is extremely tedious and involves an inordinate amount of algebraic manipulation, we omit these details, though we did test the correctness of our implementation against the outputs of \verb|JAX|'s autodifferentiation.

\subsection{Acknowledgments}
We thank Lizhi (Gary) Yang for help setting up the hardware experiments. We thank Victor Dorobantu for discussions that led to a valid formulation and proof of Theorem \ref{thm:containment}. We thank Thomas Lew for discussions regarding Hausdorff distance bounds that inspired the condition of Theorem \ref{thm:containment}. We thank Georgia Gkioxari for discussions about object representations.

Finally, we thank the maintainers of all the open-source software used extensively in this work, including but not limited to \verb|Drake|, \verb|Pinocchio|, \verb|torchquad|, \verb|JAX|, \verb|sdfstudio|, \verb|trimesh|, \verb|nlopt|, \verb|VHACD|, and \verb|quantecon| \citeapp{gomez2021_torchquad, jax2018_github, yu2022_sdfstudio, trimesh, johnson2011_nlopt, kraft1988_slsqp}.

\bibliographystyleapp{unsrt}
\bibliographyapp{references}

\end{document}

%% file: notation.tex
\newcommand{\pdf}{p} 
\newcommand{\pos}{x} 
\newcommand{\posSet}{\mathcal{X}}

%% file: iros_rewrite_cameraready_arxiv.bbl
\begin{thebibliography}{1}

\bibitem{amos2017_diffopt}
Brandon Amos and J.~Zico Kolter.
\newblock \href{https://arxiv.org/abs/1703.00443}{OptNet: Differentiable
  Optimization as a Layer in Neural Networks}.
\newblock In {\em International Conference on Machine Learning}, 2017.

\bibitem{gomez2021_torchquad}
Pablo Gómez, Håvard~Hem Toftevaag, and Gabriele Meoni.
\newblock torchquad: Numerical integration in arbitrary dimensions with
  pytorch.
\newblock {\em Journal of Open Source Software}, 6(64):3439, 2021.

\bibitem{jax2018_github}
James Bradbury, Roy Frostig, Peter Hawkins, Matthew~James Johnson, Chris Leary,
  Dougal Maclaurin, George Necula, Adam Paszke, Jake Vander{P}las, Skye
  Wanderman-{M}ilne, and Qiao Zhang.
\newblock \href{https://jax.readthedocs.io/en/latest/index.html}{{JAX}:
  composable transformations of {P}ython+{N}um{P}y programs}, 2018.

\bibitem{yu2022_sdfstudio}
Zehao Yu, Anpei Chen, Bozidar Antic, Songyou Peng, Apratim Bhattacharyya,
  Michael Niemeyer, Siyu Tang, Torsten Sattler, and Andreas Geiger.
\newblock Sdfstudio: A unified framework for surface reconstruction, 2022.

\bibitem{trimesh}
{Dawson-Haggerty et al.}
\newblock \href{https://github.com/mikedh/trimesh}{trimesh}, 2019.

\bibitem{johnson2011_nlopt}
Steven~G. Johnson.
\newblock \href{http://ab-initio.mit.edu/nlopt}{The NLopt
  nonlinear-optimization package}, 2011.

\bibitem{kraft1988_slsqp}
Dieter Kraft.
\newblock
  \href{http://degenerateconic.com/uploads/2018/03/DFVLR_FB_88_28.pdf}{A
  software package for sequential quadratic programming}.
\newblock {\em Forschungsbericht Deutsche Forschungs und Versuchsanstalt fur
  Luft und Raumfahrt}, 1988.

\end{thebibliography}


\begin{thebibliography}{10}

\bibitem{roa2014_graspmetricssurvey}
M{\'a}ximo~A. Roa and Ra{\'u}l Su{\'a}rez.
\newblock
  \href{https://link.springer.com/article/10.1007/s10514-014-9402-3}{Grasp
  quality measures: review and performance}.
\newblock {\em Autonomous Robots}, 38:65 -- 88, 2014.

\bibitem{li1988_taskorientedgrasping}
Z.~Li and S.S. Sastry.
\newblock Task-oriented optimal grasping by multifingered robot hands.
\newblock {\em IEEE Journal on Robotics and Automation}, 4(1):32--44, 1988.

\bibitem{kappler2015_bigdatagrasping}
Daniel Kappler, Jeannette Bohg, and Stefan Schaal.
\newblock \href{https://ieeexplore.ieee.org/document/7139793}{Leveraging big
  data for grasp planning}.
\newblock In {\em 2015 IEEE International Conference on Robotics and Automation
  (ICRA)}, pages 4304--4311, 2015.

\bibitem{shao2019_unigrasp}
Lin Shao, F{\'a}bio Ferreira, Mikael Jorda, Varun Nambiar, Jianlan Luo, Eugen
  Solowjow, Juan~Aparicio Ojea, Oussama Khatib, and Jeannette Bohg.
\newblock \href{https://arxiv.org/abs/1910.10900}{UniGrasp: Learning a Unified
  Model to Grasp with N-Fingered Robotic Hands}.
\newblock {\em ArXiv}, abs/1910.10900, 2019.

\bibitem{ferraricanny1992}
C.~Ferrari and J.~Canny.
\newblock \href{https://ieeexplore.ieee.org/document/219918}{Planning optimal
  grasps}.
\newblock In {\em Proceedings 1992 IEEE International Conference on Robotics
  and Automation}, pages 2290--2295 vol.3, 1992.

\bibitem{li2023_frogger}
Albert~H. Li, Preston Culbertson, Joel~W. Burdick, and Aaron~D. Ames.
\newblock \href{https://arxiv.org/abs/2302.13687}{FRoGGeR: Fast Robust Grasp
  Generation via the Min-Weight Metric}.
\newblock {\em ArXiv}, abs/2302.13687, 2023.

\bibitem{weisz2012_pfc_pose_robust}
Jonathan Weisz and Peter~K. Allen.
\newblock \href{https://ieeexplore.ieee.org/document/6224697}{Pose error robust
  grasping from contact wrench space metrics}.
\newblock In {\em 2012 IEEE International Conference on Robotics and
  Automation}, pages 557--562, 2012.

\bibitem{dragiev2011_gpis}
Stanimir Dragiev, Marc Toussaint, and Michael Gienger.
\newblock \href{https://ieeexplore.ieee.org/document/5980395}{Gaussian process
  implicit surfaces for shape estimation and grasping}.
\newblock In {\em 2011 IEEE International Conference on Robotics and
  Automation}, pages 2845--2850, 2011.

\bibitem{mahler2015_gpis}
Jeffrey Mahler, Sachin Patil, Ben Kehoe, Jur van~den Berg, Matei Ciocarlie,
  Pieter Abbeel, and Ken Goldberg.
\newblock \href{https://ieeexplore.ieee.org/document/7139882}{GP-GPIS-OPT:
  Grasp planning with shape uncertainty using Gaussian process implicit
  surfaces and Sequential Convex Programming}.
\newblock In {\em 2015 IEEE International Conference on Robotics and Automation
  (ICRA)}, pages 4919--4926, 2015.

\bibitem{defarias_tactile_gpis_uncertain}
Cristiana de~Farias, Naresh Marturi, Rustam Stolkin, and Yasemin Bekiroglu.
\newblock \href{https://arxiv.org/abs/2103.00655}{Simultaneous Tactile
  Exploration and Grasp Refinement for Unknown Objects}.
\newblock {\em IEEE Robotics and Automation Letters}, PP, 02 2021.

\bibitem{li2016_shape_uncertainty}
Miao Li, Kaiyu Hang, Danica Kragic, and Aude Billard.
\newblock
  \href{https://www.sciencedirect.com/science/article/pii/S0921889015001967}{Dexterous
  grasping under shape uncertainty}.
\newblock {\em Robotics and Autonomous Systems}, 75:352--364, 2016.

\bibitem{mahler2017_dexnet2}
Jeffrey Mahler, Jacky Liang, Sherdil Niyaz, Michael Laskey, Richard Doan, Xinyu
  Liu, Juan~Aparicio Ojea, and Ken Goldberg.
\newblock \href{https://arxiv.org/abs/1703.09312}{Dex-Net 2.0: Deep Learning to
  Plan Robust Grasps with Synthetic Point Clouds and Analytic Grasp Metrics}.
\newblock {\em ArXiv}, abs/1703.09312, 2017.

\bibitem{lu2018_planningmultifinger}
Qingkai Lu, Kautilya Chenna, Balakumar Sundaralingam, and Tucker Hermans.
\newblock \href{https://arxiv.org/abs/1804.03289}{Planning Multi-Fingered
  Grasps as Probabilistic Inference in a Learned Deep Network}.
\newblock In {\em International Symposium of Robotics Research}, 2018.

\bibitem{lu2020_diffgrasplearning}
Qingkai Lu, Mark Van~der Merwe, Balakumar Sundaralingam, and Tucker Hermans.
\newblock \href{https://arxiv.org/abs/2001.09242}{Multifingered Grasp Planning
  via Inference in Deep Neural Networks: Outperforming Sampling by Learning
  Differentiable Models}.
\newblock {\em IEEE Robotics and Automation Magazine}, 27(2):55--65, 2020.

\bibitem{liu2020_deepdiffgrasp}
Min Liu, Zherong Pan, Kai Xu, Kanishka Ganguly, and Dinesh Manocha.
\newblock \href{https://arxiv.org/abs/2002.01530}{Deep Differentiable Grasp
  Planner for High-DOF Grippers}.
\newblock {\em ArXiv}, abs/2002.01530, 2020.

\bibitem{liu2021_diversediffgrasps}
Tengyu Liu, Zeyu Liu, Ziyuan Jiao, Yixin Zhu, and Song-Chun Zhu.
\newblock \href{https://arxiv.org/abs/2104.09194}{Synthesizing Diverse and
  Physically Stable Grasps With Arbitrary Hand Structures Using Differentiable
  Force Closure Estimator}.
\newblock {\em IEEE Robotics and Automation Letters}, 7:470--477, 2021.

\bibitem{wang2022_dexgraspnet}
Ruicheng Wang, Jialiang Zhang, Jiayi Chen, Yinzhen Xu, Puhao Li, Tengyu Liu,
  and He~Wang.
\newblock \href{https://arxiv.org/abs/2210.02697}{DexGraspNet: A Large-Scale
  Robotic Dexterous Grasp Dataset for General Objects Based on Simulation}.
\newblock {\em ArXiv}, abs/2210.02697, 2022.

\bibitem{wu2022_learningdexgraspsgenmodel}
Albert Wu, Michelle Guo, and C.~Karen Liu.
\newblock \href{https://arxiv.org/abs/2207.00195}{Learning Diverse and
  Physically Feasible Dexterous Grasps with Generative Model and Bilevel
  Optimization}.
\newblock {\em ArXiv}, abs/2207.00195, 2022.

\bibitem{pokorny2013_lowerboundfc}
Florian~T. Pokorny and Danica Kragic.
\newblock \href{https://ieeexplore.ieee.org/document/6696854}{Classical grasp
  quality evaluation: New algorithms and theory}.
\newblock In {\em 2013 IEEE/RSJ International Conference on Intelligent Robots
  and Systems}, pages 3493--3500, 2013.

\bibitem{dai2015_forceclosuresdp}
Hongkai Dai, Anirudha Majumdar, and Russ Tedrake.
\newblock
  \href{https://link.springer.com/chapter/10.1007/978-3-319-51532-8_18}{Synthesis
  and Optimization of Force Closure Grasps via Sequential Semidefinite
  Programming}.
\newblock In {\em International Symposium of Robotics Research}, 2015.

\bibitem{mildenhall2020_nerf}
Ben Mildenhall, Pratul~P. Srinivasan, Matthew Tancik, Jonathan~T. Barron, Ravi
  Ramamoorthi, and Ren Ng.
\newblock \href{https://arxiv.org/abs/2003.08934}{NeRF: Representing Scenes as
  Neural Radiance Fields for View Synthesis}.
\newblock {\em CoRR}, abs/2003.08934, 2020.

\bibitem{hayashi2017_bivariategaussianintegral}
Naoki Hayashi, Kohei Segawa, and Shigemasa Takai.
\newblock \href{https://www.tandfonline.com/doi/abs/10.9746/jcmsi.10.110}{2D
  Voronoi Coverage Control with Gaussian Density Functions by Line
  Integration}.
\newblock {\em SICE Journal of Control, Measurement, and System Integration},
  10(2):110--116, 2017.

\bibitem{oneill2006_diffgeom}
B.~O'Neill.
\newblock {\em Elementary Differential Geometry, Revised 2nd Edition}.
\newblock Elsevier Science, 2006.

\end{thebibliography}
